%% file: egpaper_for_review.tex
\documentclass[10pt,twocolumn,letterpaper]{article}

\usepackage{iccv}
\usepackage{times}
\usepackage{epsfig}
\usepackage{graphicx}
\usepackage{amsmath}
\usepackage{amssymb}
\usepackage{longtable}
\usepackage{afterpage}
\usepackage{magicabbrev}

\usepackage{siunitx}
\DeclareSIUnit\px{px}
\usepackage{glossaries}
\include{magicacronyms}

\glsdisablehyper
\defglsentryfmt{
	\ifglsused{\glslabel}{
		\glsgenentryfmt
	}{
	\emph{\glsgenentryfmt}}}
\newcommand{\Tstrut}{} 
\newcommand{\Bstrut}{}

\input{figures.tex}

\input{tables.tex}

\newcommand{\skyscapes}{SkyScapes }

\newif\ifnotes
\notestrue

\usepackage{arydshln}
\usepackage{rotating}
\usepackage{contour}
\usepackage{amssymb}
\usepackage{subcaption}

\usepackage[pagebackref=true,breaklinks=true,letterpaper=true,colorlinks,bookmarks=false]{hyperref}
\usepackage{cleveref}

\iccvfinalcopy 

\def\iccvPaperID{3556} 

\ificcvfinal\fi

\begin{document}
\title{SkyScapes -- Fine-Grained Semantic Understanding of Aerial Scenes}


\newcommand\Mark[1]{\textsuperscript#1}
\author{ 
Seyed Majid Azimi\Mark{1}
\quad
Corentin Henry\Mark{1}
\quad
Lars Sommer\Mark{2}
\quad
Arne Schumann\Mark{2}
\quad
Eleonora Vig\Mark{1}\\
\\
\Mark{1}{German Aerospace Center (DLR), Wessling, Germany}
\qquad
\Mark{2}{Fraunhofer IOSB, Karlsruhe, Germany}\\
{\tt\small
\url{https://www.dlr.de/eoc/en/desktopdefault.aspx/tabid-12760}}\\ 
{Corresponding author: seyedmajid.azimi@dlr.de}
}


\makeatletter
\let\@oldmaketitle\@maketitle
\renewcommand{\@maketitle}{\@oldmaketitle
 \vspace{-.5cm}
  \includegraphics[width=\linewidth,height=8\baselineskip]
    {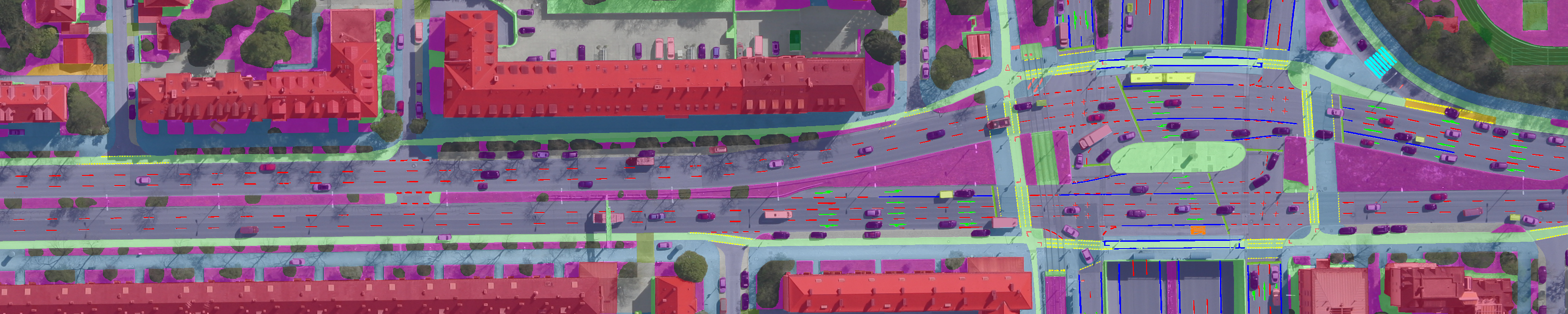}\\
    \centering
    Aerial image with overlaid annotation: dense (19 classes) and lane markings (12 classes); the dataset covers $5.7~km^{2}$.
    \bigskip}
\makeatother
\maketitle
\thispagestyle{empty} 
\input{sections/abstract}
\input{sections/intro}
\input{sections/dataset}
\input{sections/methods}

\input{sections/eval}
\input{sections/conclusion}

\input{sections/acknowledgment}
{\small
\bibliographystyle{ieee_fullname}
\bibliography{egpaper_for_review}
}
\end{document}


\title{SkyScapes -- Fine-Grained Semantic Understanding of Aerial Scenes
\\ -- Supplementary Material -- 
}


\newcommand\Mark[1]{\textsuperscript#1}
\author{ 
Seyed Majid Azimi\Mark{1}
\quad
Corentin Henry\Mark{1}
\quad
Lars Sommer\Mark{2}
\quad
Arne Schumann\Mark{2}
\quad
Eleonora Vig\Mark{1}\\
\\
\Mark{1}{German Aerospace Center (DLR), Wessling, Germany}
\qquad
\Mark{2}{Fraunhofer IOSB, Karlsruhe, Germany}\\
{\tt\small
\url{https://www.dlr.de/eoc/en/desktopdefault.aspx/tabid-12760}} 
}

\maketitle

\input{sections/supplementary}
 {\small
 \bibliographystyle{ieee_fullname}
 \bibliography{egbib}
 }

%% file: magicacronyms.tex
\newacronym[]{gls:RS}{RS}{remote sensing}
\newacronym[]{gls:GSD}{GSD}{ground sampling distance}
\newacronym[]{gls:CNN}{CNN}{convolutional neural network}
\newacronym[]{gls:FCNs}{FCNs}{fully-convolutional neural networks}
\newacronym[]{gls:CRF}{CRF}{conditional random field}
\newacronym[]{gls:SGD}{SGD}{Stochastic Gradient Descent}
\newacronym[]{gls:FCNN}{FCNN}{fully convolutional neural network}
\newacronym[]{gls:DCNN}{DCNN}{deep convolutional neural network}
\newacronym[]{gls:ACNN}{ACNN}{atrous convolutional neural network}
\newacronym[longplural={oriented bounding boxes}]{gls:OBB}{OBB}{oriented bounding box}
\newacronym[longplural={horizontal bounding boxes}]{gls:HBB}{HBB}{horizontal bounding box}
\newacronym[]{gls:RCNN}{RCNN}{Region-based convolutional neural network}

\newacronym[]{gls:miou}{mIoU}{mean Intersection over Union}

\newacronym[]{AV}{AV}{autonomous vehicles}
\newacronym[]{HD}{HD}{high definition}
\newacronym[]{ADAS}{ADAS}{advanced driver assistance systems}
\newacronym[]{GPS}{GPS}{global positioning system}
\newacronym[]{USGS}{USGS}{U.S. Geological Survey}
\newacronym[]{DWT}{DWT}{discrete wavelet transform}
\newacronym[]{SVM}{SVM}{support vector machine}
\newacronym[]{DSM}{DSM}{digital surface model}
\newacronym[]{IU}{IoU}{Intersection over Union}

\newglossaryentry{gls:FastRCNN}{name={Fast-\gls{gls:RCNN}}, description={}}
\newglossaryentry{gls:FasterRCNN}{name={Faster-\gls{gls:RCNN}}, description={}}
\newglossaryentry{gls:MaskRCNN}{name={Mask-\gls{gls:RCNN}}, description={}}

\newacronym[]{cml}{cm-level}{centimeter-level}
\newacronym[]{UAV}{UAV}{Unmanned Aerial Vehicle}
\newacronym[]{FDB}{FDB}{fully dense block}
\newacronym[]{FRSR}{FRSR}{full-resolution separable residual}
\newacronym[]{RB}{RB}{residual block}
\newacronym[]{CRASPP}{CRASPP}{concatenated reverse ASPP}
\newacronym[]{ASPP}{ASPP}{atrous spatial pyramid pooling block}
\newacronym[]{LKBR}{LKBR}{large-kernels with boundary refinement}

%% file: figures.tex
\newcommand{\samplebigImage}{%
\begin{figure*}
\centering
\begin{subfigure}{\textwidth}
\includegraphics[width=\linewidth]{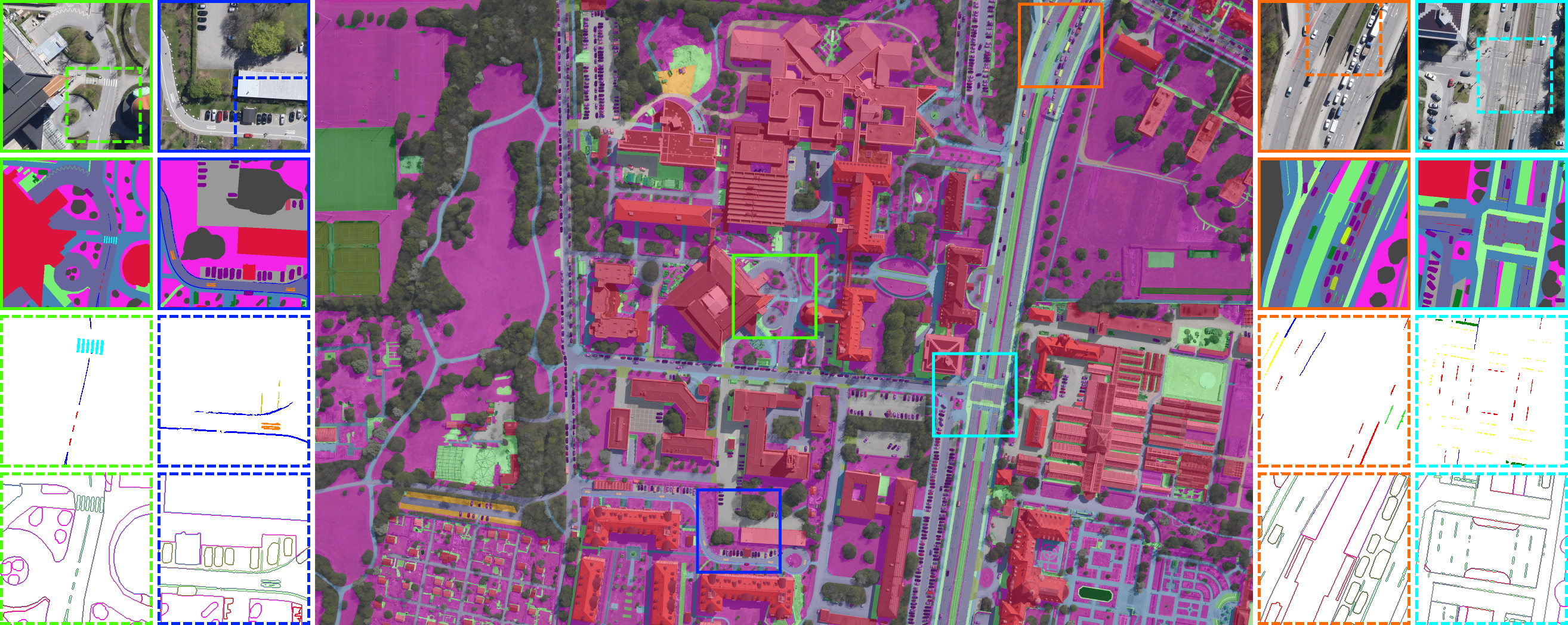}
\end{subfigure}
\caption{SkyScapes image with overlaid annotation and zoomed-in samples ($\times2$: solid line, $\times4$: dashed line). Top to bottom: RGB, dense annotation (20 classes), lane markings annotation (12 classes), multi-class edges. Class colors as in~\cref{fig:stats_skyscape}.}
\label{fig:sampleGTSkyscapes}
\end{figure*}
}

\newcommand{\benchmarkFigDense}{%
\begin{figure*}
    \centering
    \begin{subfigure}[t]{0.196\linewidth}
        \includegraphics[width=\linewidth]{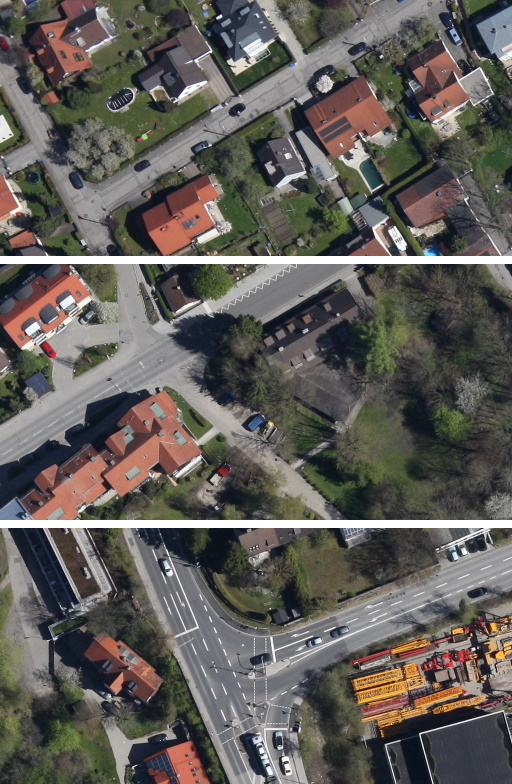}
        \caption{RGB Image}
    \end{subfigure}
    \begin{subfigure}[t]{0.196\linewidth}
        \includegraphics[width=\linewidth]{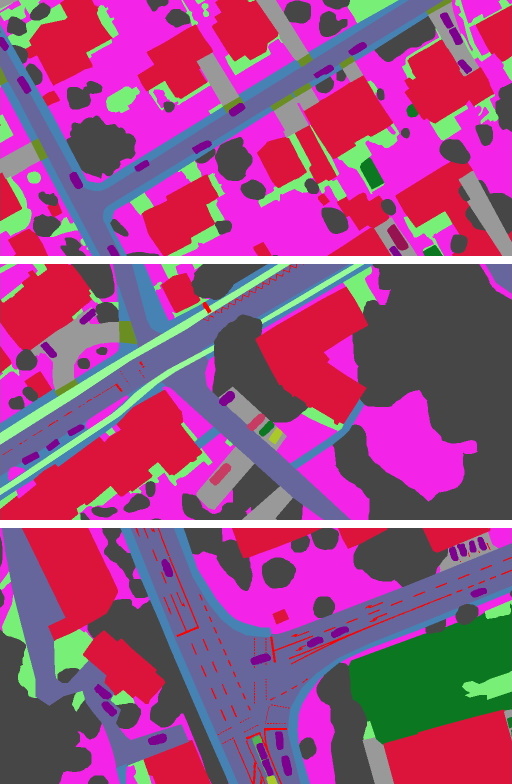}
        \caption{Ground Truth}
    \end{subfigure}
    \begin{subfigure}[t]{0.196\linewidth}
        \includegraphics[width=\linewidth]{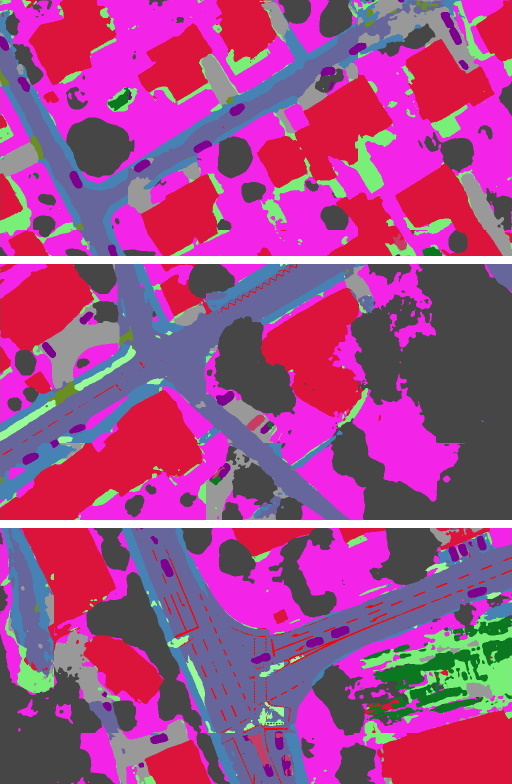}
        \caption{SkyScapesNet}
    \end{subfigure}
    \begin{subfigure}[t]{0.196\linewidth}
        \includegraphics[width=\linewidth]{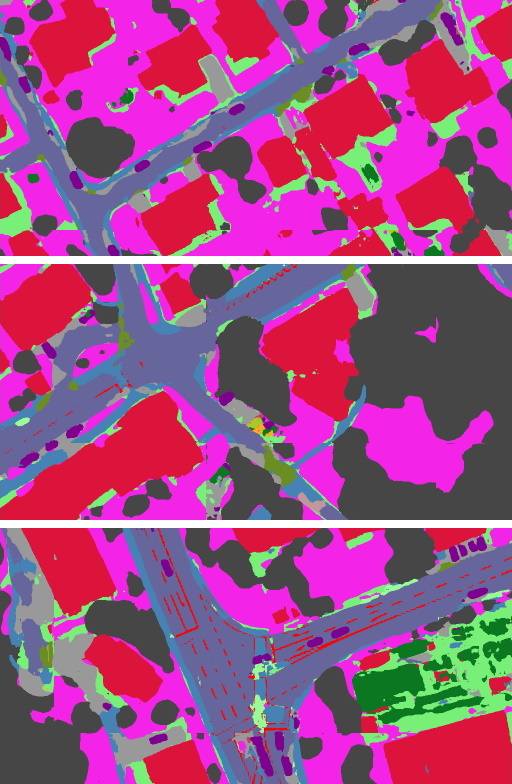}
        \caption{DeepLabv3+}
    \end{subfigure}
    \begin{subfigure}[t]{0.196\linewidth}
        \includegraphics[width=\linewidth]{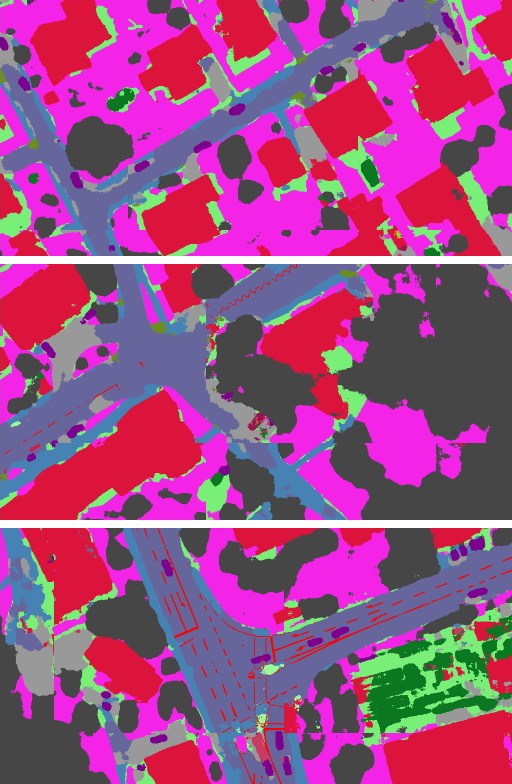}
        \caption{FC-DenseNet103}
    \end{subfigure}
    \caption{Result samples for SkyScapes-Dense task by SkyScapesNet and the two best baselines. For class colors, cf.\ \cref{fig:stats_skyscape}.}
    \label{fig:QRdense}
    \vspace{-0.2cm}
\end{figure*}
}

\newcommand{\benchmarkFigLane}{%
\begin{figure}
    \centering
    \begin{subfigure}[t]{0.24\linewidth}
        \includegraphics[width=\linewidth]{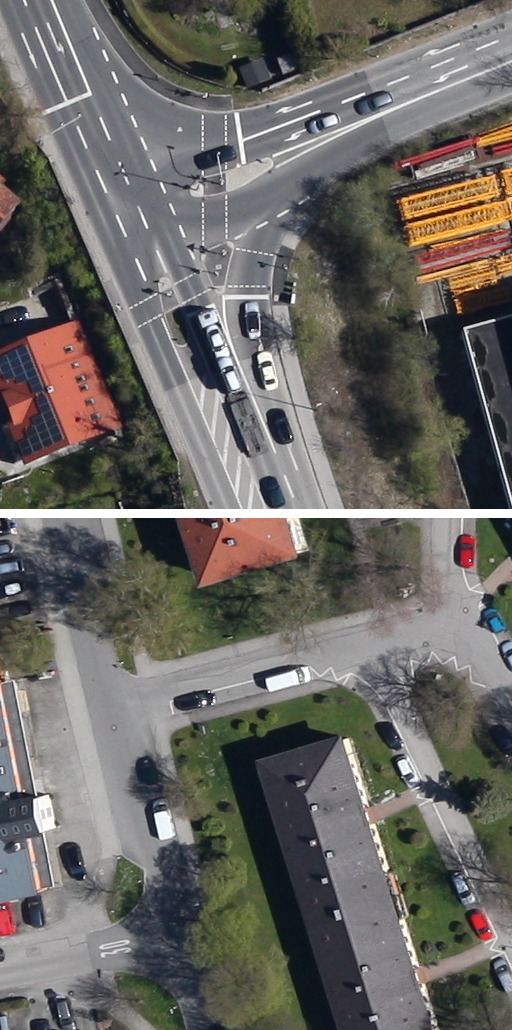}
        \caption{Image}
    \end{subfigure}
    \begin{subfigure}[t]{0.24\linewidth}
        \includegraphics[width=\linewidth]{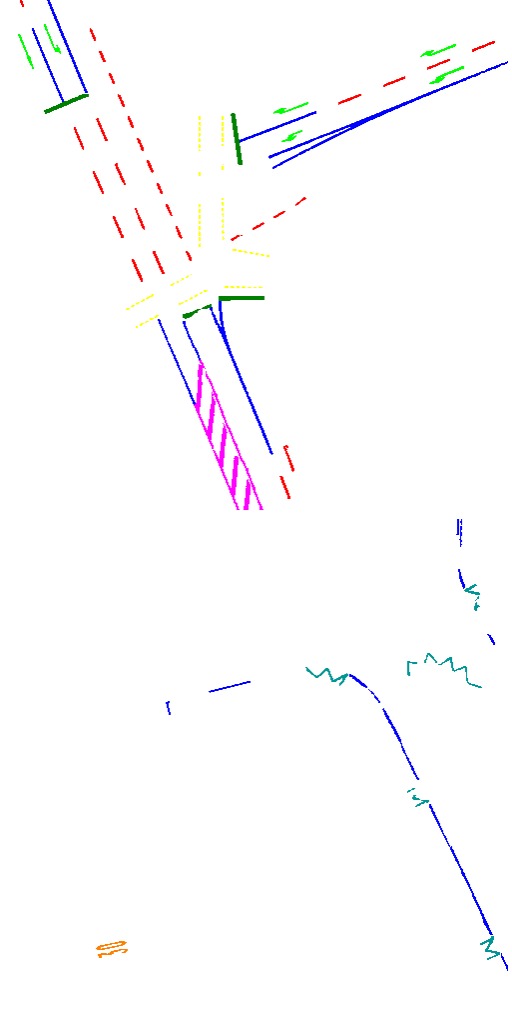}
        \caption{GT}
    \end{subfigure}
    \begin{subfigure}[t]{0.24\linewidth}
        \includegraphics[width=\linewidth]{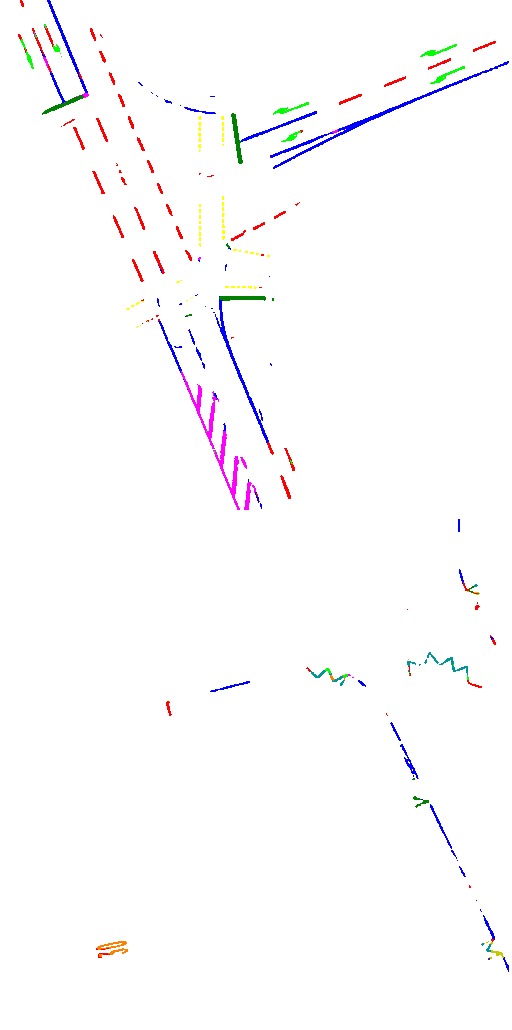}
        \caption{Ours}
    \end{subfigure}
    \begin{subfigure}[t]{0.24\linewidth}
        \includegraphics[width=\linewidth]{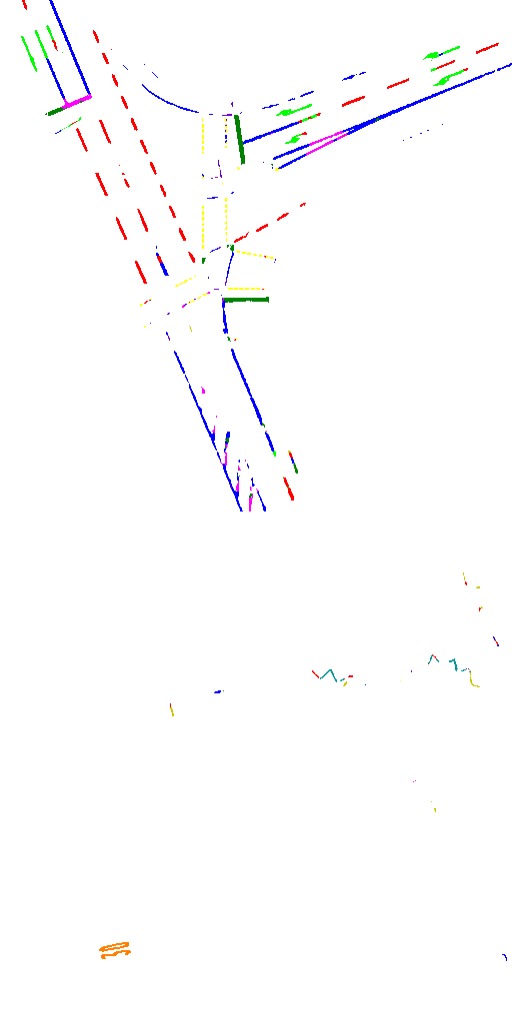}
        \caption{Enc-Dec*}
        \end{subfigure}
    \caption{Result samples for the SkyScapes-Lane  task by SkyScapesNet and the best baseline. Class colors: cf.\ \cref{fig:stats_skyscape}.}
    \label{fig:QRlane}
\end{figure}
}



%% file: tables.tex
\newcommand\fst[1]{\underline{\textbf{#1}}} 
\newcommand\snd[1]{\textbf{#1}} 

\newcommand{\experimentalresultsSkyScapesDense}{%
\begin{table}[t]
	\centering
		 \caption{Benchmark of the state of the art on the SkyScapes-Dense task over all 20 classes; `-' means no specific back-bone; `f.w.' is frequency weighted IoU; * skip connections.}
		\resizebox{0.48\textwidth}{!}{
		\addtolength{\tabcolsep}{-2pt}
		\begin{tabular}{@{\hspace{0pt}}ccccccccccc@{}}
		\multicolumn{1}{c}{Method} & 
		\multicolumn{1}{c}{Base} & 
		\multicolumn{2}{c}{IoU $[\%]$} &
		\multicolumn{2}{c}{average  $[\%]$}\\
        \multicolumn{1}{c}{} & 
		\multicolumn{1}{c}{} & 
        \multicolumn{1}{c}{mean} & 
        \multicolumn{1}{c}{f.w.} & 
        \multicolumn{1}{c}{recall} & 
        \multicolumn{1}{c}{prec.}\Bstrut\\
	    \hline
        FCN-8s~\cite{long2015fully}         & ResNet50 & 33.06 &67.02 &40.78 &\fst{65.01}\Tstrut\\
        SegNet~\cite{badrinarayanan2017segnet}         & --       & 23.14 &61.32 &29.21 &59.56 \Tstrut\\
        U-Net~\cite{ronneberger2015unet}           & --       & 14.15 &36.33 &21.88 &22.87\Tstrut\\
        BiSeNet~\cite{yu2018bisenet}& ResNet50 & 30.82 &59.62 &40.25 &49.42 \Tstrut\\
        DenseASPP~\cite{yang2018denseaspp}      & ResNet101& 24.73 &56.58 &32.21 &40.82 \Tstrut\\
        Encoder-Decoder*&   --     & 37.16 &67.18 &\fst{48.26} &50.16 \Tstrut\\
        FC-DenseNet-103~\cite{jegou2017one}  &   --     & 37.78 &67.44 &46.66 &53.89 \Tstrut\\
        FRRNA~\cite{pohlen2017fullresolution}          &   --     & 37.20 &65.10 &46.44 &53.22 \Tstrut\\
        GCN~\cite{peng2017large}            & ResNet152& 32.92 &65.12 &41.60 &49.65 \Tstrut\\
        Mobile-U-Net* & --       & 34.96 &65.26 &44.52 &49.49 \Tstrut\\
        PSPNet~\cite{zhao2017pyramid}         & ResNet101& 30.44 &61.62 &40.48 &43.63 \Tstrut\\
        RefineNet~\cite{lin2017refinenet}      & ResNet152& 36.39 &65.52 &46.12 &52.17 \Tstrut\\
        DeepLabv3+~\cite{chen2018deeplab}    & Xception65 & \snd{38.20}&\snd{68.81}&47.97&55.34\Tstrut\\
        \hline
        SkyScapesNet           &     --   & \fst{40.13} &\fst{72.67} &\snd{47.85} & \snd{65.93} \Tstrut\\
	\end{tabular}
	
	\vspace{-0.2cm}
	}\label{tab:bchmkSkyscapesAllclsFinal}
\end{table}
}

\newcommand{\experimentalresultsSkyScapesLane}{%
\begin{table}[t]
	\centering
		 \caption{Benchmark of the state of the art on the SkyScapes-Lane task over all 13 classes. Cf.\ table \ref{tab:bchmkSkyscapesAllclsFinal} for abbreviations.}
		\resizebox{0.49\textwidth}{!}{
		\addtolength{\tabcolsep}{-2pt}
		\begin{tabular}{@{\hspace{0pt}}ccccccccccc@{}}
		\multicolumn{1}{c}{Method} & 
		\multicolumn{1}{c}{Base} & 
		\multicolumn{2}{c}{IoU $[\%]$} &
		\multicolumn{2}{c}{average  $[\%]$}\\
        \multicolumn{1}{c}{} & 
		\multicolumn{1}{c}{} & 
        \multicolumn{1}{c}{mean} & 
        \multicolumn{1}{c}{f.w.} & 
        \multicolumn{1}{c}{recall} & 
        \multicolumn{1}{c}{precision}\Bstrut\\
	 \hline
        FCN-8s~\cite{long2015fully} & ResNet50 & 13.74 &99.69 & 15.23 & 77.96 \Tstrut\\ 
        U-Net~\cite{ronneberger2015unet} & -- & 8.97 &99.62 &12.73 &\fst{88.26}\Tstrut\\
        AdapNet~\cite{valada2017adaptnet} & -- & 20.20 &99.67 &22.21 &53.60\Tstrut\\
        BiSeNet~\cite{yu2018bisenet} & ResNet50 & 23.77 &99.66 &28.71 &51.42\Tstrut\\
        DeepLabv3~\cite{chen2017rethinking} & ResNet50 & 16.15 &99.62 &18.94 &55.44\Tstrut\\
        DenseASPP~\cite{yang2018denseaspp} & ResNet101 & 17.00 &99.65 &18.74 &46.02\Tstrut\\
        FC-DenseNet-103~\cite{jegou2017one} & -- & 48.42 &99.85 &\snd{55.32}&69.01\Tstrut\\
        FRRN-B~\cite{pohlen2017fullresolution}  & -- & 47.02 &99.85 &54.72 &66.19\Tstrut\\
        GCN~\cite{peng2017large} & ResNet50 & 35.65 &99.82 &43.09 &55.65\Tstrut\\
        Mobile-U-Net* & -- & 41.21 &99.84 &47.48 &64.60\Tstrut\\
        PSPNet~\cite{zhao2017pyramid} & ResNet101 & 35.85 &99.82 &42.64 &58.23\Tstrut\\
        DeepLabv3+~\cite{chen2018deeplab} & Xception65 & 37.14 &99.77 &43.14 &62.07\Tstrut\\
        Encoder-Decoder*  & -- & \snd{48.87} &\snd{99.85} &55.31 &70.63\Tstrut\\
        \hline
        SkyScapesNet & -- & \fst{51.93} &\fst{99.87} &\fst{60.53} &\snd{72.29} \Tstrut\\
        
	\end{tabular}
	\vspace{-0.2cm}
	}\label{tab:bchmkSkyscapesMultiLaneAllcls}
\end{table}
}

\newcommand{\experimentalresultsSkyScapesCatgory}{%
\begin{table}[t]
	\centering
		 \caption{Results on SkyScapes-Dense-Category, multi-class edge, and binary edge prediction tasks.}
		\resizebox{0.48\textwidth}{!}{
		\addtolength{\tabcolsep}{-2pt}
		\begin{tabular}{@{\hspace{0pt}}ccccccccccc@{}}
		\multicolumn{1}{c}{Method} & 
		\multicolumn{1}{c}{Task} & 
		\multicolumn{2}{c}{IoU $[\%]$} &
		\multicolumn{2}{c}{average  $[\%]$}\\
        \multicolumn{1}{c}{} & 
		\multicolumn{1}{c}{} & 
        \multicolumn{1}{c}{mean} & 
        \multicolumn{1}{c}{f.w.} & 
        \multicolumn{1}{c}{recall} & 
        \multicolumn{1}{c}{prec.}\Bstrut\\
	    \hline
        SkyScapesNet    &     Category           & 52.27 &77.77 &63.49 &65.65 \Tstrut\\ 
        SkyScapesNet    &     Multi-class Edge   & 13.00 &88.74 &16.82 &22.74 \Tstrut\\ 
        SkyScapesNet     &    Binary Edge        & 58.72 &89.52 &64.81 &71.99\Tstrut\\ 
         
	\end{tabular}
	\vspace{-0.4cm}
	}\label{tab:bchmkSkyscapesCategory}
\end{table}
}

\newcommand{\ablationstudy}{%
\begin{table}
    \centering
    \caption{Evaluation of different parts of SkyScapesNet. `Baseline' was trained only with cross-entropy (\ie no IoU loss added). 
    Max stride is 32\,pixels. * using original number of sub-sampling as in the baseline in SkyScapesNet.
    }
	\resizebox{0.39\textwidth}{!}{

	\begin{tabular}{@{\hspace{0pt}}c|c|c|c|c|c|c|c|c|cc@{}}
	\rotatebox[origin=c]{90}{Network}&

	\multicolumn{1}{c}{\rotatebox[origin=c]{90}{loss IoU}} & 
	\multicolumn{1}{c}{\rotatebox[origin=c]{90}{sep. branch.}} & 
	\multicolumn{1}{c}{\rotatebox[origin=c]{90}{FDB}} & 
	\multicolumn{1}{c}{\rotatebox[origin=c]{90}{FSRRB}} & 
	\multicolumn{1}{c}{\rotatebox[origin=c]{90}{CRASPP}} & 
	\multicolumn{1}{c}{\rotatebox[origin=c]{90}{LKBR}} & 
	\multicolumn{1}{c}{\rotatebox[origin=c]{90}{mIoU $[\%]$}}\\
    \hline
    Baseline*~\cite{jegou2017one} &                   & & & & & & 37.78\\
    Baseline &                   & & & & & & 36.88\\
    SkyScapesNet &\checkmark&           & & & & & 37.08\\
    SkyScapesNet &\checkmark&\checkmark&          &           &           &           & 38.55\\
    SkyScapesNet &\checkmark&\checkmark&\checkmark&           &           &           & 38.77\\
    SkyScapesNet &\checkmark&\checkmark&\checkmark&\checkmark &           &           & 38.90\\
    SkyScapesNet &\checkmark&\checkmark&\checkmark&\checkmark &\checkmark &           & 39.09\\
    SkyScapesNet &\checkmark&\checkmark&\checkmark&\checkmark &\checkmark &\checkmark & 39.30\\
    SkyScapesNet* &\checkmark&\checkmark&\checkmark&\checkmark &\checkmark &\checkmark & \textbf{40.13}\\

    \end{tabular}
    \vspace{-0.2cm}
    }\label{tab:ablation}
\end{table}
}

\newcommand{\postdamhoustonresults}{%
\begin{table}[t]
	\centering
		 \caption{Generalization of our model trained on SkyScapes-Dense and evaluated on Potsdam and DFC2018.}
		\resizebox{0.49\textwidth}{!}{
		\addtolength{\tabcolsep}{-2pt}
		\begin{tabular}{@{\hspace{0pt}}ccccccccccc@{}}
		\multicolumn{1}{c}{training data} & 
		\multicolumn{1}{c}{test data} & 
		\multicolumn{2}{c}{IoU $[\%]$} &
		\multicolumn{2}{c}{average  $[\%]$}\\
        \multicolumn{1}{c}{} & 
		\multicolumn{1}{c}{} & 
        \multicolumn{1}{c}{mean} & 
        \multicolumn{1}{c}{f.w.} & 
        \multicolumn{1}{c}{recall} & 
        \multicolumn{1}{c}{prec.}\Bstrut\\
	    \hline
        SkyScapes   &  Potsdam & 47.46 &70.58 &62.28 &66.09\Tstrut\\ 
        SkyScapes   &  Data Fusion Contest 2018 & 26.42 & 47.58 & 55.67 &37.64 & \Tstrut\\ 

	\end{tabular}
	\vspace{-0.2cm}
	}\label{tab:bchmkSkyscapesAllclsAux}
\end{table}
}



%% file: sections/abstract.tex
\begin{abstract}
\vspace{-0.25cm}
   Understanding the complex urban infrastructure with centimeter-level accuracy is essential for many  applications from autonomous driving  to mapping, infrastructure monitoring, and urban management. Aerial images provide valuable information over a large area instantaneously; nevertheless, no current dataset captures the complexity of aerial scenes at the level of granularity required by real-world applications. To address this, we introduce SkyScapes, an aerial image dataset with highly-accurate, fine-grained annotations for pixel-level semantic labeling. SkyScapes provides annotations for 31 semantic categories ranging from large structures, such as buildings, roads and vegetation, to fine details, such as 12 (sub-)categories of lane markings. We have defined two main tasks on this dataset: dense semantic segmentation and multi-class lane-marking prediction. We carry out extensive experiments to evaluate state-of-the-art segmentation methods on SkyScapes. 
   Existing methods struggle to deal with the wide range of classes, object sizes, scales,  and fine details present. We therefore propose a novel multi-task model, which incorporates semantic edge detection and is better tuned for feature extraction from a wide range of scales. This model achieves notable improvements over the baselines in region outlines and level of detail on both tasks. 
   \vspace{-0.25cm}
\end{abstract}

%% file: sections/intro.tex
\section{Introduction}
\label{sec:intro}
Automated methods for creating maps of today's urban and rural infrastructures with \gls{cml} accuracy are of great aid in handling their growing complexity.
Applications of such accurate maps include urban management, city planning, and infrastructure monitoring/maintenance.
Another prominent example is the creation of \gls{HD} maps for autonomous driving.
Applications here include the use of a general road network for navigation and more advanced automation tasks in \Gls{ADAS}, such as lane departure warnings, which rely on precise information about lane boundaries, sidewalks, 
etc.~\cite{poggenhans2018lanelet2,seif2016autonomous,mattyus2016hd,zheng2018high,mattyus2017deeproadmapper}.

Currently, the data collection process to generate \gls{HD} maps is mainly carried out by so-called mobile mapping systems, which comprise of a vehicle equipped with a broad range of sensors (\eg Radar, LiDAR, cameras) followed by automated analysis of the collected data ~\cite{guo2016low,gwon2017generation,carneiro2018mapping,lamon2006mapping}.
The limited field-of-view and occlusions due to the oblique sensor angle make this automated analysis complicated. 
In addition, mapping large urban areas in this way requires a lot of time and resources.
An aerial perspective can alleviate many of these problems and simultaneously allow for processing of much larger areas of \gls{cml} geo-referenced data in a short time.
Existing aerial semantic segmentation datasets, however, are limited in the range of their annotations. They either focus on a few individual classes, such as roads or building footprints in the INRIA~\cite{maggiori2017dataset}, Massachusetts~\cite{mnih2013thesis}, SpaceNet~\cite{van2018spacenet}, or DeepGlobe~\cite{demir2018deepglobe} datasets, or they provide very coarse classes, such as the GRSS\_DFC\_2018~\cite{le20182018}, or the ISPRS Vaihingen and Potsdam datasets~\cite{potsdam}. Other datasets are recorded at  sensor angles and at flight heights unsuitable for \gls{HD} mapping~\cite{lyu2018uavid,semanticdrone} or contain potentially inaccurate annotations generated automatically~\cite{wang2016torontocity}.
In addition, only few works tackle lane-marking extraction in aerial imagery, and they either rely on third-party sources such as OpenStreetMap, or only provide a binary extraction in Azimi et al.~\cite{azimi2018aerial}.

Ground imagery has greatly benefited from large-scale datasets, such as ImageNet~\cite{deng2009imagenet}, Pascal VOC~\cite{everingham2010pascal}, MS-COCO~\cite{lin2014microsoft}, but in aerial imagery the annotation is scarce and more tedious to obtain. 
%
In this work, we propose a new aerial image dataset, called SkyScapes, which closes this gap by providing detailed annotations of urban scenes for established classes, such as buildings, vegetation, and roads, as well as fine-grained classes, such as various types of lane markings, vehicle entrance/exit zones, danger areas, etc. 
Fig.\ \ref{fig:sampleGTSkyscapes} shows sample  annotations offered by SkyScapes. 

The dataset contains 31 classes and a rigorous annotation process was established to provide a high degree of annotation accuracy.
\skyscapes uniquely combines the fine-grained annotation of road infrastructure with an overhead viewing angle and coverage of large areas, thus enabling the generation of \gls{HD} maps for various applications.
We evaluate several state-of-the-art semantic segmentation models as baselines on SkyScapes.
Existing models achieve a significantly lower accuracy on our dataset than on established benchmarks with either ground-views or a much coarser set of classes.
Our analysis of the most common errors hints at many merged regions and inaccurate boundaries.
We therefore propose a novel segmentation model, which incorporates semantic edge detection as an auxiliary task.
The secondary loss function emphasizes edges more strongly during the learning process, leading to a clear reduction of the prominent error cases.
Furthermore, the proposed architecture takes both large- and small-scale objects into account. 

In summary:
i) we provide a new aerial dataset for semantic segmentation with highly accurate annotations and fine-grained classes, thus enabling the development of models for previously unsupported tasks, such as aerial HD-mapping;
ii) we carry out extensive evaluations of current state-of-the-art models and show that existing approaches struggle to handle the large number of classes and level of detail in the dataset;
iii) hence, we propose a new multi-task model, which combines semantic segmentation with edge detection, yielding more precise region outlines.

%% file: sections/dataset.tex
\section{The \skyscapes Dataset}
\label{sec:properties}
\samplebigImage
The data collection was carried out with a helicopter flying over the greater area of Munich, Germany. 
A low-cost camera system~\cite{kurz2011real,gstaiger2015airborne}  consisting of three standard DSLR cameras and 
mounted on a flexible platform was used for recording the data, with only the nadir-looking capturing images. 
In total, 16 non-overlapping RGB images of size \SI{5616x3744}\,pixels were  chosen. The flight altitude of about \SI{1000}{\m} above ground led to a \gls{gls:GSD} of approximately \SI{13}{\cm/pixel}. The images represent urban and partly rural areas with highways, first/second order roads, and complex traffic situations, such as crossings and congestion, as exemplified in~\cref{fig:sampleGTSkyscapes}.

\subsection{Classes and Annotations}
Thirty-one semantic categories were  annotated: low vegetation, paved road, non-paved road, paved parking place, non-paved parking place, bike-way, sidewalk, entrance/exit, danger area, building, car, trailer, van, truck, large truck, bus, clutter, impervious surface, tree, and 12 lane-marking types. The considered lane-markings are the following: dash-line, long-line, small dash-line, turn sign, plus sign, other signs, crosswalk, stop-line, zebra zone, no parking zone, parking zone, other lane-markings.
The selection of classes was influenced by their relevance to real-world applications, hence, road-like objects dominate.  
Class definitions and visual examples for each class are given in the supplementary materials, class statistics can be found in Fig.~\ref{fig:stats_skyscape}.

The SkyScapes dataset was manually annotated using tools adapted to each object class and following a strict annotation policy.
Annotating aerial images requires considerable time and effort, especially when dealing with many small objects, such as lane-markings. Shadows, occlusion, and unclear object boundaries also add to the difficulty.
Due to the size and shape complexity, and to the large number of classes/instances, annotation required considerably more work than for ground-view benchmarks (such as CityScapes~\cite{cordts2016cityscapes}), also limiting the dataset size. 
To ensure high quality, the annotation process was performed iteratively with a three-level quality check over each class, overall taking about 200 man-hours per image.
We show one such annotated image in Fig.~\ref{fig:sampleGTSkyscapes}.

In SkyScapes, we enforce pixel-accurate annotations, as even small offsets lead to large localization errors in aerial images (\eg a 1-pixel offset in SkyScapes would lead to a 13\,cm error). As autonomous vehicles require a min.\ accuracy of 20\,cm for on-map localization \cite{Ziegler2014BerthaDrive}, we chose the highly accurate annotation of a smaller set of images over coarser annotations of a much larger set.  
%
In fact, in~\cref{subsec:generalization}, we show high generalization of our model when trained on \skyscapes and tested on third-party data. 

\subsection{Dataset Splits and Tasks}
\label{subsec:splitsTasks}
We split the dataset into training, validation, and test sets with 50\%, 12.5\%, and 37.5\% portions respectively. We chose this particular split due to the class imbalance and to avoid splitting larger images. 


Lane-markings and the rest of the scene elements (such as buildings, roads, vegetation, and vehicles) present different challenges, with lane-markings operating on much finer scales and requiring a fine-grained differentiation, whereas other scene elements are represented on a much wider scale. 
Having considered these challenges, we defined five different tasks: 1) \textbf{SkyScapes-Dense} with 20 classes as the lane-markings were merged into a single class, 2) \textbf{SkyScapes-Lane} with 13 classes comprising 12 lane-marking classes and a non-lane-marking one, 3) \textbf{SkyScapes-Dense-Category} with 11 merged classes comprising nature (low-vegetation, tree), driving-area (paved, non-paved), parking-area (paved, non-paved), human-area (bikeway, sidewalk, danger area), shared human and vehicle area (entrance/exit), road-feature (lane-marking), residential area (building), dynamic-vehicle (car, van, truck, large-truck, bus), static-vehicle (trailer), man-made surface (impervious surface), and others objects (clutter), 4) \textbf{SkyScapes-Dense-Edge-Binary}, and 5) \textbf{SkyScapes-Dense-Edge-Multi}. The two latter tasks are binary and multi-class edge detection, respectively. 
Defining separate tasks allows for more fine-grained control to fit the model to the dense object regions, their boundaries, and their classes. This is especially helpful when object boundary accuracy is paramount and difficult to extract, \eg for multi-class lane-markings.


\subsection{Statistical Properties}

\begin{figure}[ht]
    \centering
		\begin{subfigure}[b]{\linewidth}
    	\caption{SkyScapes-Dense}
    	\vspace{-0.15cm}
    	\includegraphics[trim=10 10 5 11, clip, width=\linewidth]{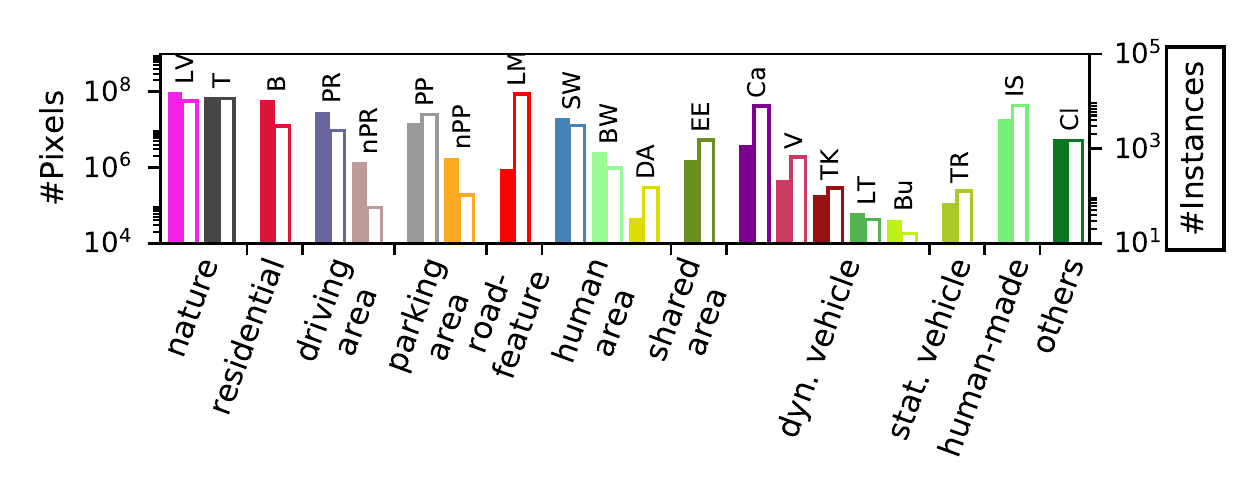}
	\end{subfigure}
	\begin{subfigure}[b]{\linewidth}
	    \caption{SkyScapes-Lane}
	    \vspace{-0.15cm}
  	    \includegraphics[trim=10 10 5 11, clip,width=\linewidth]{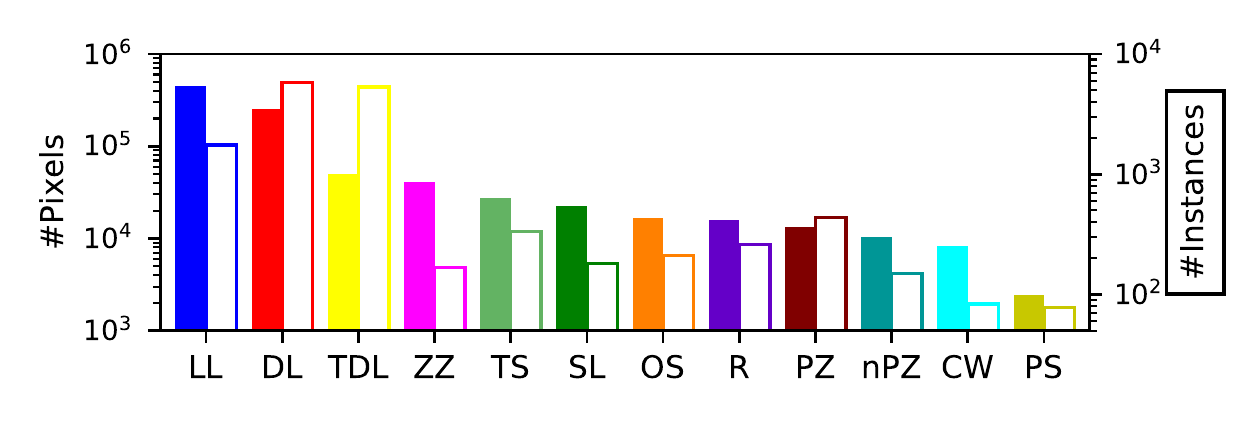}
	\end{subfigure}
    \caption{Number of annotated pixels (filled) and instances (non-filled) per class in  SkyScapes-Dense and SkyScapes-Lane for  low-vegetation (LV), tree (T), building (B), paved-road (PR), paved-parking-place (PP), non-paved-parking-place (nPP), non-paved-road (nPR), lane-marking (LM), sidewalk (SW), bikeway (BW), danger-area (DA), entrance-exit (EE), car (Ca), van (V), truck (TK), trailer (TR), long-truck (LT), bus (Bu), impervious-surface (IS), clutter (Cl), long line (LL), dash line (DL), tiny dash line (TDL), zebra zone (ZZ), turn sign (TS), stop line (SL), other signs (OS), the rest of lane-markings (R), parking zone (PZ), no parking zone (nPZ), crosswalk (CW), and plus sign (PS).}
    \label{fig:stats_skyscape}
\end{figure}



SkyScapes is comprised of more than 70K annotated instances that are divided into 31 classes. The number of annotated pixels and instances per class for SkyScapes-Dense and SkyScapes-Lane are given in \cref{fig:stats_skyscape}. 
The majority of pixels are annotated as low vegetation, tree, or building, whereas the most common classes are lane markings, tree, low vegetation, and car. This illustrates the wide range from classes with fewer large regions to those with many small regions.
A similar range can be observed among the lane markings within the more fine-grained SkyScapes-Lane task.
With an average pixel area of about 9 pixels, `tiny dash lines' are the smallest instances. 

A quantitative comparison of SkyScapes against existing aerial segmentation datasets is provided in~\cref{tab:stats}.
Existing datasets lack the high detail level and annotation quality of SkyScapes. Potsdam contains fewer classes (6 vs 31), less accurate labels, and image distortions due to suboptimal orthorectification. TorontoCity focuses on quantity: its wider spatial coverage requires (a less precise) automated labeling.
SkyScapes offers the largest number of classes including various fine-structures (\eg lane markings). In absolute terms, SkyScapes contains also notably more region instances, which emphasizes the higher complexity of SkyScapes. Handling this range of classes and variety of object instance sizes is one of the main challenges. The capability of state-of-the-art segmentation methods to address these challenges has not yet been thoroughly explored.

\begin{table*}
    \begin{center}
    \caption{Statistics of SkyScapes and  other aerial datasets. To date, TorontoCity is not publicly available.}
        \vspace{0.2cm}
    	\resizebox{.85\textwidth}{!}{
        \begin{tabular}{lccccc}
         & SkyScapes & Potsdam~\cite{potsdam} & Vaihingen~\cite{potsdam} & Aerial KITTI~\cite{mattyus2015enhancing} & TorontoCity~\cite{wang2016torontocity} \\
        \hline
        Classes & 31 & 6 & 6 & 4 & 2+8 \\
        Images & 16 & 38 & 33 & 20 & N/A\\
        Image dimension (px) & 5616$\times$3744  & 6000$\times$6000 & 2493$\times$2063 (avg) & variable & N/A \\
        GSD (cm/pixel) & 13 & 5 & 9 & 9 & 10 \\
        Aerial coverage (km$^2$) & 5.69 (urban\&rural) & 3.42 & 1.36 & 3.23 & 712 \\
        Instances & 70,346  & 42,389 & 10,700 & 2,814 & N/A
        \end{tabular}
        }
    \label{tab:stats}
    \end{center}
    \vspace{-0.5cm}

\end{table*}

\section{Semantic Benchmarks} 
In the following, we review several state-of-the-art segmentation methods and benchmark these on SkyScapes.

 \subsection{Metrics}
To assess the segmentation performance, we use the Jaccard Index, known as the PASCAL VOC \gls{IU} metric:  $\frac{TP}{TP+FP+FN}$~\cite{everingham2010pascal}, where TP, FP, and FN stand  for the numbers of true positive, false positive, and false negative pixels for each class, determined over the test set. We also report other metrics, such as frequency weighted IoU, pixel accuracy, average recall/precision, and mean IoU, \ie the average of IoUs over all classes as defined in~\cite{long2015fully}. In the supplementary material, we report $IoU_{class}$ for SkyScapes-Dense and $IoU_{category}$ for the best baseline on SkyScapes-Dense-Category. 
Unlike in the street scenes of CityScapes~\cite{cordts2016cityscapes}, in aerial scenes the objects can be as long as the image size (roads or long-line lane-markings). Therefore, we do not report $IoU_{instance}$. 

\subsection{State of the Art in Semantic Segmentation}

As detection results have matured, reaching around 80\% mean AP on Pascal VOC~\cite{kim2018parallel} and on the DOTA aerial object detection dataset~\cite{xia2018dota,azimi2018towards}, 
the interest has shifted to pixel-level segmentation,  which yields a more detailed localization of an object and handles occlusion better than bounding boxes. In recent years, \gls{gls:FCNs}~\cite{long2015fully,sermanet2013overfeat}  achieved remarkable performance on several semantic segmentation benchmarks. Current state-of-the-art methods include Auto-Deeplab~\cite{liu2019autodeeplab}, DenseASPP~\cite{yang2018denseaspp}, BiSeNet~\cite{yu2018bisenet}, Context-Encoding~\cite{zhang2018context}, and OcNet~\cite{yuan2018ocnet}. While specific architecture choices offer a good baseline performance, the integration of a multi-scale context aggregation module is key to competitive performance. Indeed, context information is crucial in pixel labeling tasks. It is best leveraged by so-called ``pyramid pooling modules'', using either stacks of input images at different scales, as in PSPNet~\cite{zhao2017pyramid}, or stacks of convolutional layers with different dilation rates, as in DeepLab~\cite{chen2014semantic}.
However, context aggregation is often performed at the expense of fine-grained details. As a remedy, FRRN~\cite{pohlen2017fullresolution} implements an architecture comprising a full-resolution stream for segmenting the details and a separate pooling stream for analyzing the context. Similarly, GridNet~\cite{fourure2017gridnet} uses multiple interconnected streams working at several resolutions.
For our benchmark, in addition to the aforementioned models, we train several other popular segmentation networks: FCN~\cite{long2015fully}, U-Net~\cite{ronneberger2015unet}, MobileNet~\cite{howard2017mobilenets}, SegNet~\cite{badrinarayanan2017segnet}, RefineNet~\cite{lin2017refinenet}, Deeplabv3+~\cite{chen2018encoder}, AdapNet~\cite{valada2017adaptnet}, and FC-DenseNet~\cite{jegou2017one}, as well as a custom U-Net-like MobileNet and custom Decoder-Encoder with skip-connections. 

In tables~\ref{tab:bchmkSkyscapesAllclsFinal} and \ref{tab:bchmkSkyscapesMultiLaneAllcls}, we report our benchmarking results for the above methods.
As anticipated, all methods struggle on \skyscapes due to the significant differences between ground and aerial imagery exposed in the introduction. On the SkyScapes-Dense task (table~\ref{tab:bchmkSkyscapesAllclsFinal}), classification mistakes are for the most part found around the inter-class boundaries. We observe the same inter-class misclassification on the SkyScapes-Lane task (table~\ref{tab:bchmkSkyscapesMultiLaneAllcls}), and furthermore notice that many lane-markings are entirely missed and classified as background, certainly due to their few-pixel size. Both tasks hence represent a new type of challenge.
This is reinforced by the fact that the performance of the networks remained consistent from one task to the other, showing that none are specialized enough to obtain a significant advantage on either task. In our method, we tackled this challenge by focusing  on object boundaries.

%% file: sections/methods.tex
\section{Method}
\label{sec:method}
\begin{figure*}
\centering
\includegraphics[width=\linewidth]{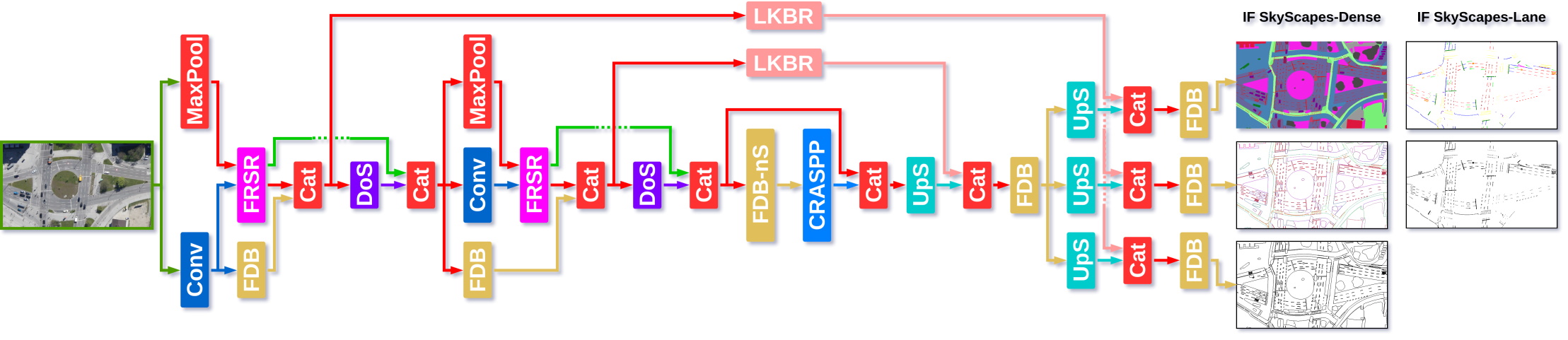}
\caption{The architecture of SkyScapesNet. Three branches are used to predict dense semantics and multi-class/binary edges. For multi-class lane-marking prediction, two branches are used to predict multi-class and binary lane-markings.}
\label{fig:wholetrimg1}
\end{figure*}

\begin{figure}
\centering
\includegraphics[width=\linewidth]{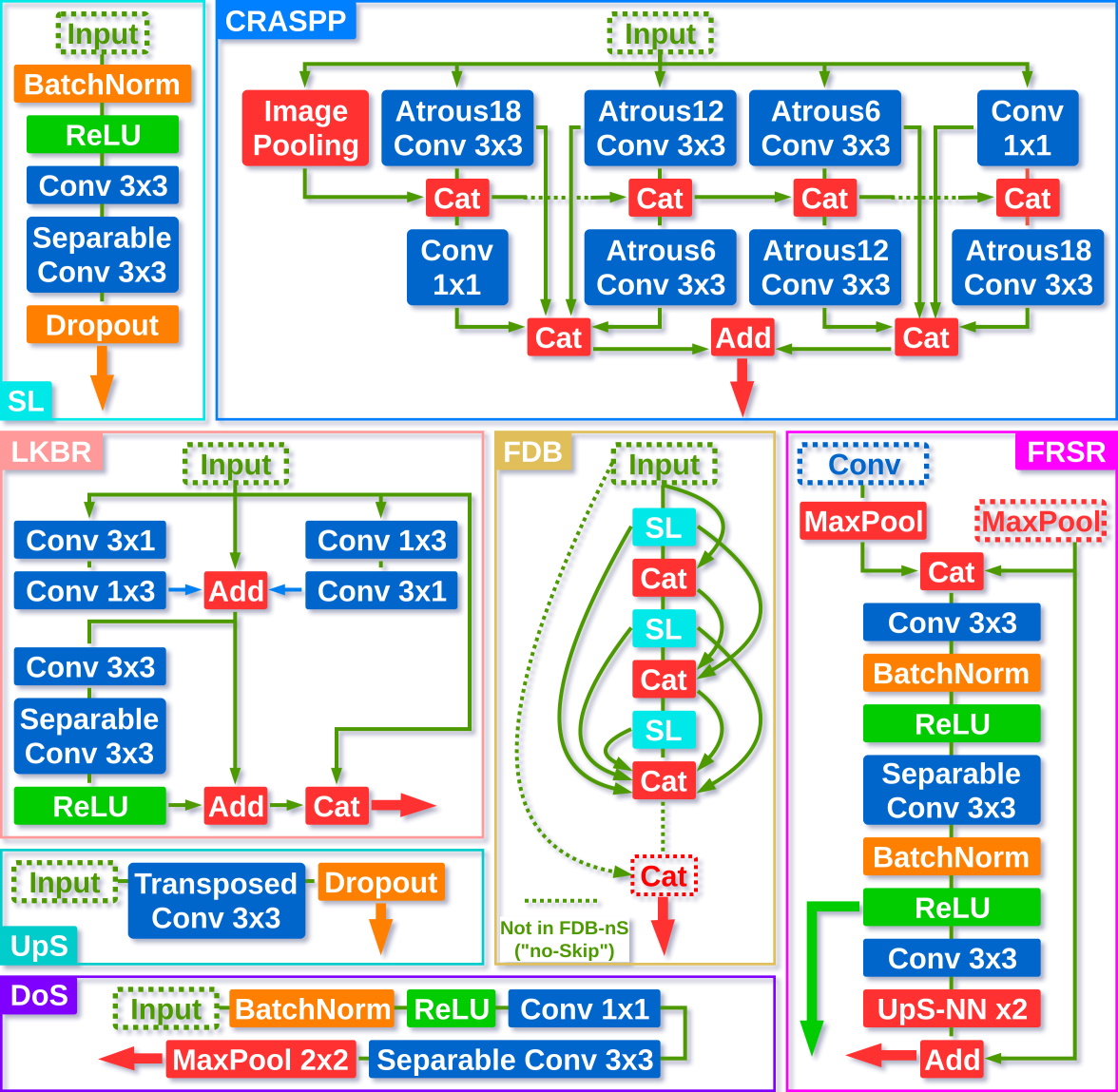}
\caption{Configuration of SkyScapesNet building blocks. SL, DoS, and UpS are Separable, Downsampling, and Upsampling blocks, UpS-NN is a Nearest-Neighbor Upsampling layer. Add/Cat are addition/concatenation operators.}
\label{fig:whole}
\end{figure}

Thirty-one highly similar classes and small complex objects in SkyScapes necessitate a specialized architecture that unifies latest architectural improvements (FC-DenseNet~\cite{jegou2017one}, auxiliary tasks, etc.) and proves more effective than the state of the art. 
Motivated by the major errors from our benchmarking analysis, we propose a multi-task method that tackles both dense prediction and edge detection to improve performance on boundary regions. In the case of multi-class lane-markings, we modify the method to enable both multi-class and binary lane-marking segmentation to decrease the number of false positives in non-lane areas. We consider FC-DenseNet~\cite{jegou2017one} as the main baseline.
SkyScapesNet, illustrated in~\cref{fig:wholetrimg1}, can be seen as a modified case of FC-DenseNet, but more generally as a multi-task ensemble-model network, encapsulating units from~\cite{jegou2017one, pohlen2017fullresolution, chen2018deeplab, peng2017large}. Thus, it also shares their advantages, such as alleviating the gradient-vanishing problem. 
\Cref{fig:whole} illustrates the building blocks, which are explained below.

\textbf{FDB:} in \gls{FDB}, we use more residual connections compared to the existing Dense Blocks (DBs) in the baseline, as inspired by DenseASPP~\cite{yang2018denseaspp}. However, instead of using atrous convolutions, we add separable-convolutions due to their recent success~\cite{chen2018deeplab}. Moreover, as SkyScapes contains large scale variation, making receptive fields larger by using larger atrous rates deteriorates the feature extraction from very small objects such as lane-markings.
The number of sub-blocks, referred to as Separable Layer (\textbf{SL}), is the same as in the DBs from the baseline.

\textbf{FRSR:} inspired by~\cite{pohlen2017fullresolution} and the comparable performance of this model with DenseNet, we add a residual-pooling stream (similar to the full-resolution residual unit -- FRRU from~\cite{pohlen2017fullresolution}) as \gls{FRSR} unit to the main stream. 
Similar to \gls{FDB}, we utilize separable convolutions. 
As the original FRRU, FRSR has two processing streams: a residual stream (for better localization) and a pooling stream (for better  recognition). Inside the pooling stream, the downsampled results go through several depth-wise separable convolutions, batch-normalization, and ReLU layers and, after applying a \SI{1x1} convolution, the output is upsampled and added to \gls{FDB}. 
We limit the number of downsamplings in \gls{FRSR} to one as the main stream applies consecutive downsampling.

\textbf{CRASPP:}  inspired by the success of \gls{ASPP}~\cite{yang2018denseaspp, chen2018encoder}, after five downsampling steps, we add the \gls{CRASPP} to enhance the feature extraction of large objects. In CRASPP, we `reverse' the original ASPP (\ie the order of atrous rates) and concatenate it with the original ASPP, so as to obtain receptive fields optimal for both small/large objects. 

 \textbf{LKBR:} for boundary refinement and to improve the extraction of tiny objects, we apply -- in addition to five skip-connections -- \glspl{LKBR}. LKBR~\cite{peng2017large} is composed of two streams 
 including a boundary refinement module.  Unlike~\cite{jegou2017one}, we apply a residual path from the output of the last downsampling module to the input of the first upsampling module.

\textbf{Multi-task learning:} we use three separate branches to predict dense semantics and multi-class and binary edges simultaneously. 
The streams are separated from each other after the second upsampling layer. The motivation is to allow the auxiliary tasks to modify the shared weights 
so as to augment the network performance on boundary regions. For multi-class lane-marking segmentation, we consider two streams with similar configuration. 

\textbf{Loss functions:}
instead of relying only on cross-entropy, we propose to add either the Soft-IoU-loss~\cite{mattyus2017deeproadmapper} or the  Soft-Dice-loss~\cite{vnet} to it (taking the sum of indiv.\ losses).  

By the direct application of the cost-aware cross-entropy loss, the network tries to fill in lane-marking areas which leads to a high TP rate for the lane-marking classes, but also high FP for the non-lane class. However, due to the very high number of non-lane pixels, the resulting FP does not have much effect on the overall accuracy. To alleviate this, we propose the scheduled weighting mechanism in which the costs of corresponding classes gradually move towards the final weighted coefficients as the training process evolves.
Further details about the architecture as well as loss formulas are included in the supplementary material.

%% file: sections/eval.tex
\section{Evaluation}
\label{sec:evaluation}
For our experiments, we crop the images into \SI{512x512} patches, as the original 21\,MP images would not fit into GPUs.
As data augmentation, we carry out horizontal and vertical flipping, and use 50\% overlap between neighboring crops both in vertical and horizontal directions. 
During inference we use 10\% overlap as a partial solution to the lower performance at  image boundaries. We use Titan XP and Quadro P6000 GPUs for training. 
The learning rate was 0.0001 and a batch size of 1 was chosen. We trained the algorithms for 60 epochs to make the comparison fair (the majority of the methods converged at this step). In total, there are 8820 training images. 
Our model has 137\,M parameters. As we deal with offline mapping, inference at 355\,ms per \SI{512x512} image patch is of little concern. 
\experimentalresultsSkyScapesDense
\benchmarkFigDense
\paragraph{SkyScapes-Dense -- 20 main classes:}
The benchmarking results reported in \cref{tab:bchmkSkyscapesAllclsFinal} demonstrate the complexity of the task. Our method described above achieves 1.93\% mIoU improvement over the best benchmark. 
Qualitative examples of the best baselines and our proposed algorithm are depicted in \cref{fig:QRdense}.
Our algorithm exhibits the best trade-off between accurately segmented coarse and fine structures.
Ablation studies in \cref{tab:ablation} quantifying the effect of several components show that the main improvement is achieved by including both binary and multi-class edge detection. 

\ablationstudy

\paragraph{SkyScapes-Lane -- multi-class lane prediction:}
Here, a further challenge is the highly imbalanced dataset.  Results in ~\cref{tab:bchmkSkyscapesMultiLaneAllcls} 
show that despite the tiny object sizes, our algorithm achieves 51.93\%\, mIoU, outperforming the state of the art by 3.06\%.
Qualitative examples in \cref{fig:QRlane} highlight that our  algorithm generates fewer decomposed segments.

\experimentalresultsSkyScapesLane
\benchmarkFigLane

\paragraph{SkyScapes-Dense -- auxiliary tasks:}
We further provide results for the three auxiliary tasks \textbf{SkyScapes-Dense-Category}, \textbf{SkyScapes-Dense-Edge-Binary}, and \textbf{SkyScapes-Dense-Edge-Multi} in \cref{tab:bchmkSkyscapesCategory} (cf.\ sec.~\ref{subsec:splitsTasks} for task definitions). 
As multiple categories are merged into a single category, \eg low vegetation and tree into nature, the mIoU for SkyScapes-Dense-Category is notably higher than for the more challenging SkyScapes-Dense.
For the edge detection branches, used to enforce the learning of more accurate boundaries, high mIoU is obtained for SkyScapes-Dense-Edge-Binary, while  still a low one for the more challenging multi-class edge detection.  

\experimentalresultsSkyScapesCatgory


\section{Generalization}
\label{subsec:generalization}

Our aim in this paper is to promote aerial imagery (in its widest sense) as a means to create HD-maps. Hence, our method is not restricted to 
aerial images captured by a helicopter, but would work for satellites and lower-flying drones, too. To demonstrate the good generalization capability of our method, here we show results on four  additional data types covering a wide range of sensors (camera and platform), spatial resolutions, and geographic locations. 

\begin{figure}
\centering
\includegraphics[width=\linewidth]{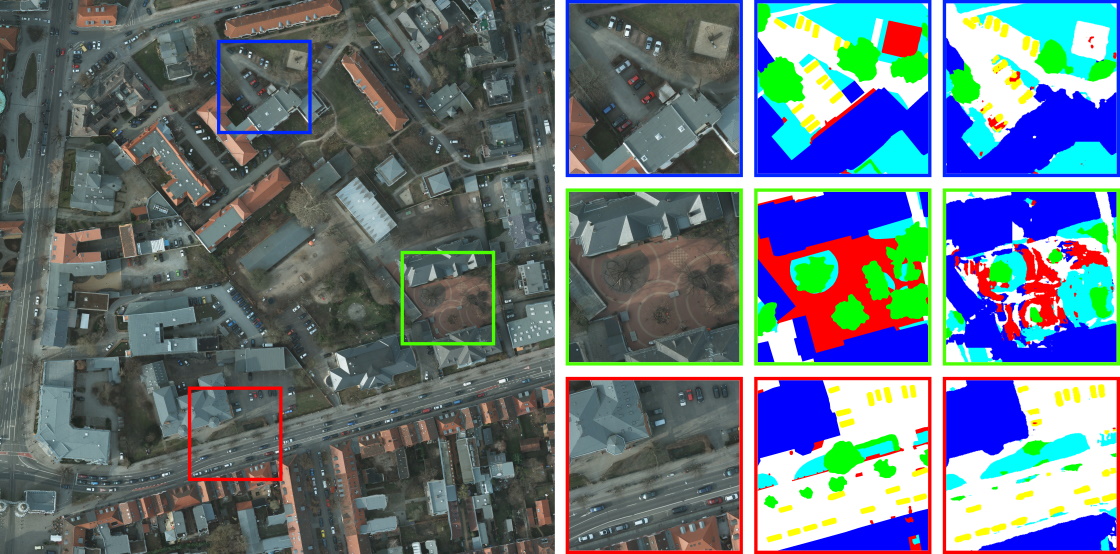}
\vspace{-0.4cm}
\caption{Results of our model trained on SkyScapes and tested on the Potsdam dataset with GSD adjustment and no fine-tuning. Patches from left to right: RGB, ground truth, prediction. Potsdam classes: 
\protect\contour{black}{\textcolor{white}{$\blacksquare$}}~impervious, \protect\contour{black}{\textcolor{blue}{$\blacksquare$}}~building, \protect\contour{black}{\textcolor{cyan}{$\blacksquare$}}~low vegetation, \protect\contour{black}{\textcolor{green}{$\blacksquare$}}~tree, \protect\contour{black}{\textcolor{yellow}{$\blacksquare$}}~car, \protect\contour{black}{\textcolor{red}{$\blacksquare$}}~clutter.}
\label{fig:benchmarkPotsdam}
\end{figure}

\begin{figure}
\centering
\includegraphics[width=\linewidth]{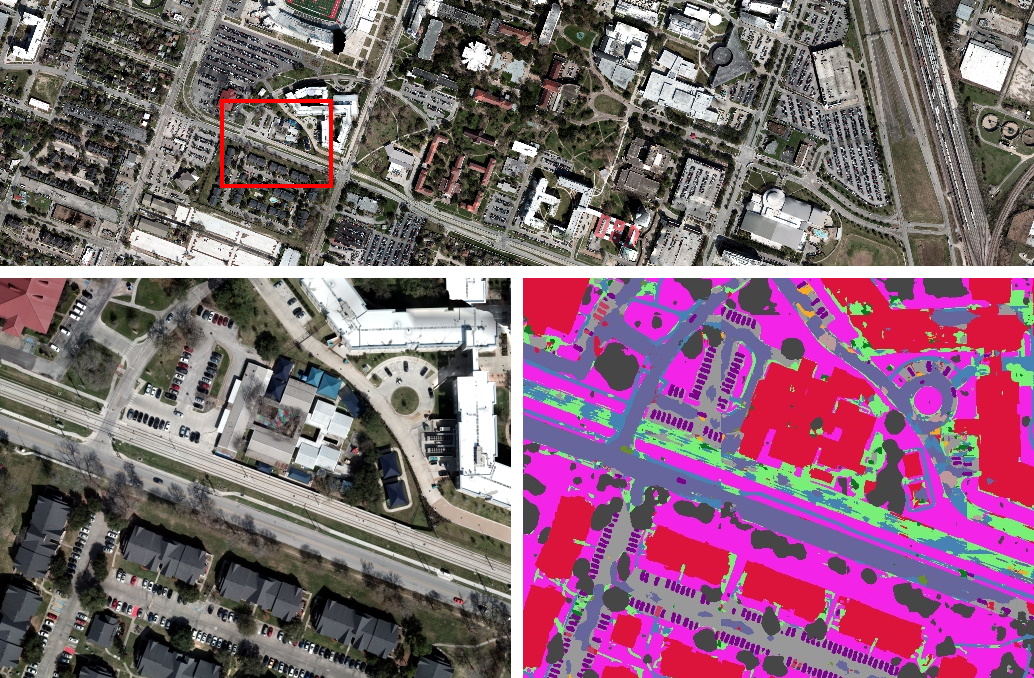}
\vspace{-0.4cm}
\caption{Results of our model trained on SkyScapes and tested on the GRSS\_DFC\_2018 dataset (over Houston, USA) with GSD adjustment and without fine-tuning.}
\label{fig:benchmarkDFCHouston}
\end{figure}

\begin{figure}
\centering
\includegraphics[width=\linewidth]{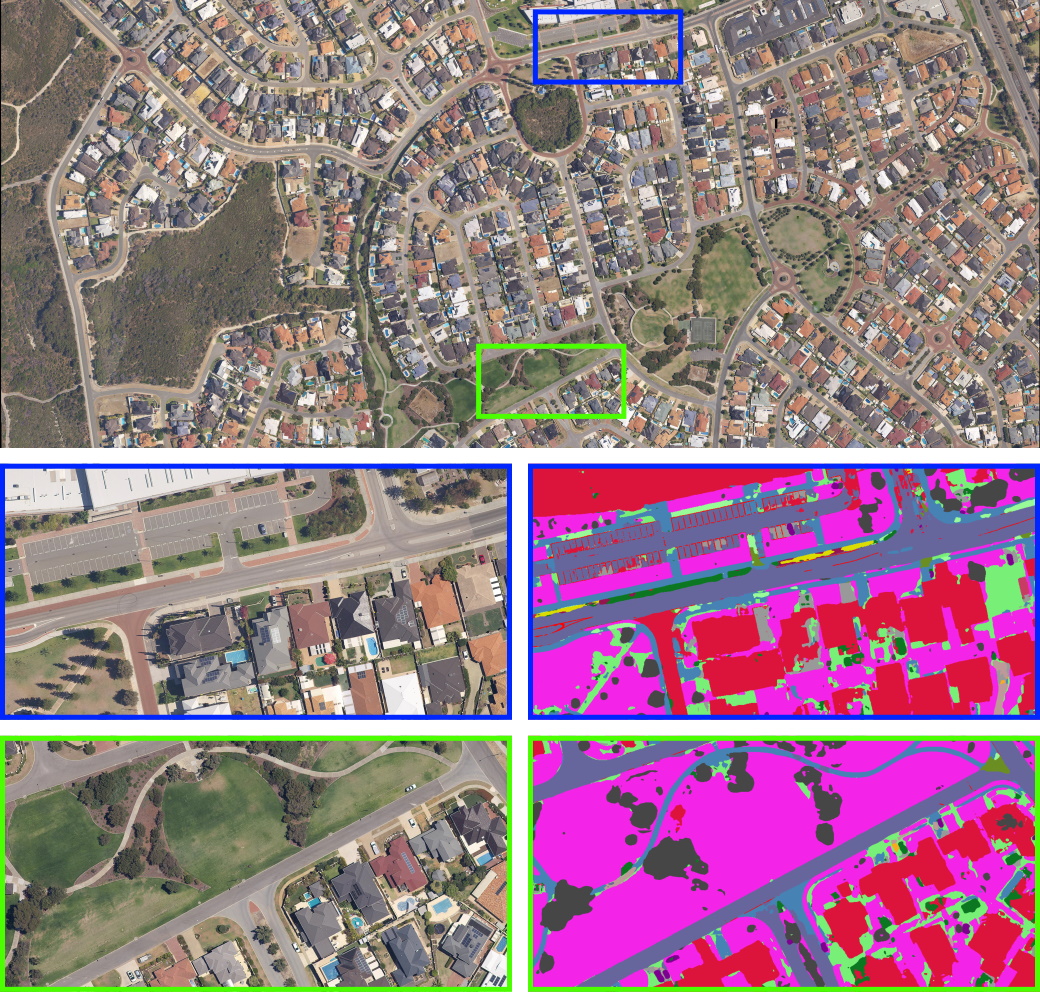}
\vspace{-0.4cm}
\caption{Segmentation result samples of our model trained on SkyScapes and tested on an aerial image over Perth, Australia, with GSD adjustment and without fine-tuning.}
\label{fig:benchmarkAustralia}
\end{figure}

For quantitative evaluation we  consider the Potsdam~\cite{potsdam} and GRSS\_DFC\_2018 datasets~\cite{le20182018}, and show qualitative results also on an aerial images of Perth, Australia.
Qualitative results can be seen in
~\cref{fig:benchmarkPotsdam,fig:benchmarkDFCHouston,fig:benchmarkAustralia}.
By adjusting the GSD of the test images (through scaling) to match that of our dataset, our model trained on SkyScapes indicates good generalization even without fine-tuning.
This is demonstrated also in the quantitative results on Potsdam (see \cref{tab:bchmkSkyscapesAllclsAux}) as the mean IoU is in the range of SkyScapes-Dense-Category.
For the quantitative evaluation, we merged our categories according to the Potsdam categories.

\postdamhoustonresults

\begin{figure}
\centering
\includegraphics[width=\linewidth]{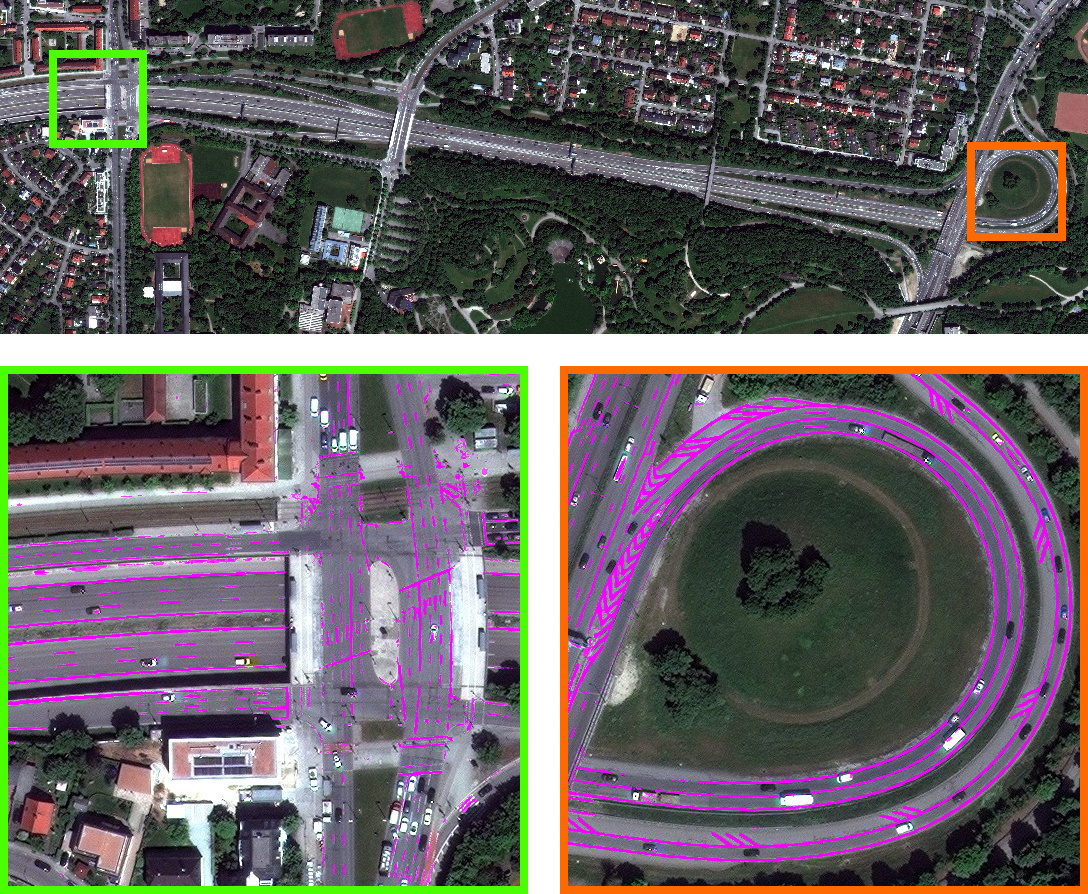}
\vspace{-0.4cm}
\caption{Binary lane segmentation on a Worldview4 satellite image over Munich using our model trained on SkyScapes, and tested on a highway scene with GSD adjustment and no fine-tuning.}
\label{fig:benchmarkWorldviewAutobahn}
\vspace{-0.5cm}
\end{figure}

Moreover, \cref{fig:benchmarkWorldviewAutobahn}
demonstrates the generalization capability of our algorithm 
for binary lane-marking extraction at a widely different scale (30\,cm/pixel) on a WorldView-4 satellite image. To the best of our knowledge, satellite images have not been used for lane-marking extraction before. 


%% file: sections/conclusion.tex
\section{Conclusion} 
In this paper, we introduced SkyScapes, an image dataset for \gls{cml} semantic labeling of aerial scenes to facilitate the creation of \gls{HD} maps for autonomous driving, urban management, city planning, and infrastructure monitoring. 
We presented an extensive evaluation of several state-of-the-art methods on SkyScapes and proposed a novel multi-task network that, thanks to its specialized architecture and auxiliary tasks, proves more effective than all tested baselines. Finally, we demonstrated good generalization of our method on four additional image types ranging from high-resolution aerial images to even satellite images. 


%% file: sections/acknowledgment.tex
\small{
\paragraph{Acknowledgements} We thank (1) Spookfish/EagleView for the aerial image over Perth; (2) the National Center for Airborne Laser Mapping and the Hyperspectral Image Analysis Lab at the Univ.\ Houston for acquiring and providing the GRSS\_DFC\_2018 data in the generalization study, the IEEE GRSS Image Analysis and Data Fusion Technical Committee; (3) Ternow AI\ GmbH for the labeling process assistance. E.\ Vig was funded by a Helmholtz Young Investigators Group grant (VH-NG-1311). 
}

%% file: sections/supplementary.tex
\section{Annotation techniques}

Several annotators worked on the creation of the ground truth, each focusing on a separate set of classes. To ensure  annotation consistency, a list of rules was established and extended as special cases were discovered. These guidelines relate to two aspects of the annotation work: target identification and boundary topology. For the former, the annotators referred to the comprehensive class definitions found in \cref{sec:supplementarySemanticClasses} to assign every object in the image to a semantic category. 
The vertical ordering of classes (or class overlays) was based on the natural physical ordering found in the real world, and as also considered in transportation systems, \ie vehicles were put on top of all road-like objects, etc. 
Some classes were annotated together to ensure that inter-object borders were not overlapping, but only after fixing the vertical class order, similarly to CityScapes~\cite{cordts2016cityscapes}: the object boundaries of low-level classes were drawn more coarsely at places where they would be overlaid with the accurate masks of higher-level classes. This sped up the annotation process while still satisfying our quality requirements.
Other objects such as vehicles were annotated separately. As a consequence, their borders did not necessarily match the boundaries of other classes in the resulting merged ground truth. In the final verification step, these seams were corrected pixel by pixel by the annotators.

\section{Semantic classes} \label{sec:supplementarySemanticClasses}
In~\cref{tab:examples}, we provide detailed definitions of the 31 annotated classes, including a typical visual example per class.

\section{Further details on SkyScapesNet}
In SkyScapesNet, we use the same number of pooling and unpooling steps as in the FC-DenseNet~\cite{jegou2017one} baseline, \ie\ 5 pooling and 5 unpooling steps. Between the encoder and decoder we use an extra \gls{FDB} module similar to the DenseBlock (DB) module in the baseline together with \gls{CRASPP}. 
The number of Separable Layers (SL) is similar to the baseline: {4, 5, 7, 10, 12, 15, 12, 10, 7, 5, 4}.
However, for the majority of the ablation studies we used the SL sequence {1, 2, 3, 4, 5, 6, 5, 4, 3, 2, 1} due to limited GPU memory in Titan XPs. The experiments marked with `*' in the ablation study table were carried out with the same number of SL modules as in the baseline.

\begin{table}
\caption{Architecture details of SkyScapesNet. The abbreviations stand for: FDB: Fully DenseBlock, DoS: Down-sampling, UpS: Up-sampling, SL: separable layer, and fm: number of feature maps. Note that skip-connections and LKBR modules have not been illustrated for simplicity.}
\resizebox{0.4\textwidth}{!}{
\begin{tabular}{ |l|l|l|l|l|l| }
\hline
\multicolumn{6}{ |c| }{\textbf{Network Architecture}} \\\hline
\multicolumn{6}{ |c| }{Input, fm=3}  \\ \hline
\multicolumn{6}{ |c| }{Convolution (3x3), fm:48} \\\hline
\multicolumn{6}{ |c| }{FDB (4 SLs), MaxPool$\rightarrow$FRSR}\\\hline
\multicolumn{6}{ |c| }{Concatenation$\rightarrow$DoS$\rightarrow$Concatenation}\\\hline
\multicolumn{6}{ |c| }{FDB (5 SLs), Conv(3x3) + MaxPool$\rightarrow$FRSR}\\\hline
\multicolumn{6}{ |c| }{Concatenation$\rightarrow$DoS$\rightarrow$Concatenation}\\\hline
\multicolumn{6}{ |c| }{FDB (7 SLs), Conv(3x3) + MaxPool$\rightarrow$FRSR}\\\hline
\multicolumn{6}{ |c| }{Concatenation$\rightarrow$DoS$\rightarrow$Concatenation}\\\hline
\multicolumn{6}{ |c| }{FDB (10 SLs), Conv(3x3) + MaxPool$\rightarrow$FRSR}\\\hline
\multicolumn{6}{ |c| }{Concatenation$\rightarrow$DoS$\rightarrow$Concatenation}\\\hline
\multicolumn{6}{ |c| }{FDB (12 SLs), Conv(3x3) + MaxPool$\rightarrow$FRSR}\\\hline
\multicolumn{6}{ |c| }{Concatenation$\rightarrow$DoS$\rightarrow$Concatenation}\\\hline
\multicolumn{6}{ |c| }{FDB (15 SLs)}\\\hline
\multicolumn{6}{ |c| }{CRASPP}\\
\hdashline
\multicolumn{6}{ |c| }{repeated in parallel for each task}\\
\hdashline
\multicolumn{6}{ |c| }{UpS + FDB (12 SLs)}\\\hline
\multicolumn{6}{ |c| }{UpS + FDB (10 SLs)}\\\hline
\multicolumn{6}{ |c| }{UpS + FDB (7 SLs)}\\\hline
\multicolumn{6}{ |c| }{UpS + FDB (5 SLs)}\\\hline
\multicolumn{6}{ |c| }{UpS + FDB (4 SLs)}\\\hline
\multicolumn{6}{ |c| }{Convolution (1x1), fm=No. of classes}\\\hline
\multicolumn{6}{ |c| }{Softmax}\\\hline
	\end{tabular}
	}\label{tab:SSNet}
\end{table}

We use HeUniform to initialize our model and train it with ADAM using a constant learning  rate  of 0.0001. We did not use any learning rate scheduler for the sake of fair benchmarking of several architectures.
We train all models  on  the augmented data with horizontal and vertical flips.
We use current batch statistics for batch normalization in all three phases: training, validation, and test.
The number of features in SL modules is the multiplication of the number of SL modules and the growth-rate. We used the same growth-rate of 16 as the baseline.
The number of feature maps in separable-convolutions is the same as in the standard convolution layers. 
We use a stride of 1 in separable convolutions.
MaxPooling is done with a kernel size of $2\times2$ with a stride of 2.
For convolutions, we use a kernel size of $3\times3$ throughout the network.
In the \gls{FRSR} module, the number of feature maps in the first convolution and in the separable convolution is twice as many as the number of feature maps in \gls{FDB} at the same step. The last convolution has equal number of feature maps as the corresponding \gls{FDB}.

The input convolution of the FRSR modules (except the first one) is $1\times1$ and the number of feature maps is equal to  $growth\;rate*number\;of\;SL\;modules$.
We use 21 feature maps in the \glspl{LKBR} modules.

In our experiments, we combine the Soft-IoU loss~\cite{mattyus2017deeproadmapper} as well as the Soft-Dice loss~\cite{vnet} with the cross-entropy loss function. 
For the multi-class segmentation task, cross-entropy is defined  as
\begin{align}\label{eq:crossentropy}
\resizebox{0.4\textwidth}{!}{$L_{cross-entropy} 
= -\frac{1}{C}\sum_{c=1}^{C}\sum_{N}{y_{nc}\log{\hat{y}_{nc}}}$}  
\end{align}
where $y_{nc} \in \{0,19\}$ is the ground-truth value for class $c$ at location $n$, $\hat{y}_{nc} \in [0,19]$ is the prediction probability, $C$ stands for the total number of classes, $N$ is the total number of pixel locations and $L$ stands for the loss function.
The Soft-IoU loss is computed as:
\begin{align}\label{eq:softiou}
\resizebox{0.4\textwidth}{!}{L_{soft-IOU} = -\frac{1}{C}\sum_{c=1}^{C}\frac{\sum_{N}{y_{nc} * \hat{y}_{nc}}}{\sum_{N}{y_{nc} + \hat{y}_{nc} - y_{nc} * \hat{y}_{nc}}}}
\end{align}
The total loss is then defined as 
\begin{align}\label{eq:softiou}
\resizebox{0.3\textwidth}{!}{L_{total} = L_{soft-IOU} + L_{cross-entropy}}
\end{align}

When the Soft-Dice loss is used, we compute the following: 
\begin{align}\label{eq:softdice}
\resizebox{0.4\textwidth}{!}{L_{soft-Dice} = -\frac{1}{C}\sum_{c=1}^{C}\frac{2 * | \sum_{N}{y_{nc} * \hat{y}_{nc}}|}{|\sum_{N}{y_{nc}}|^2 + |\sum_{N}{\hat{y}_{nc}}|^2}}
\end{align}
In~\cref{tab:ablationfull}, we evaluate the above losses on SkyScapes-Dense, both separately and in combination, and show that the combination of soft-IoU loss with cross-entropy is more beneficial than soft-Dice with cross-entropy.

\begin{table}
    \centering
    \caption{Evaluation of the different losses and their combinations on the  SkyScapes-Dense benchmark. mIoU numbers are in $[\%]$. Higher value is better. SSNet stands for SkyScapesNet.}
	\resizebox{0.45\textwidth}{!}{
	\begin{tabular}{@{\hspace{0pt}}c|c|c|c|c|@{}}
	Network&
	cross-entropy & 
	soft-IoU & 
	soft-Dice &
	mIoU $[\%]$\\
    \hline
    Baseline~\cite{jegou2017one}&\checkmark &          &          & 36.88\\
    SSNet                       &           &\checkmark&          & 36.95\\
    SSNet                       &           &          &\checkmark& 36.93\\
    SSNet                       &\checkmark &\checkmark&          & 37.08\\
    SSNet                       &\checkmark &          &\checkmark& 37.01\\

    \end{tabular}
    }\label{tab:ablationfull}
\end{table}

\section{Class merging policy for the Potsdam and GRSS\_DFC\_2018 datasets}
In order to be able to evaluate the performance of our method trained on SkyScapes on the Potsdam and GRSS\_DFC\_2018 datasets with different class definitions, we adopt the class merging policy shown in~\cref{tab:mergeforpostdam} on the SkyScapes-Dense prediction task. 
For the GRSS\_DFC\_2018 dataset, we applied a similar policy. 
\begin{table}
\caption{The class merging policy we used to make the results of our model comparable with the ground-truth labels in Potsdam.}
\resizebox{0.45\textwidth}{!}{
\begin{tabular}{ |l|l| }
\hline
SkyScapes-Dense & Potsdam \\\hline
low-vegetation & low-vegetation\\\hline
paved-road & impervious-surface\\\hline
non-paved-road & impervious-surface\\\hline
paved-parking-place & impervious-surface\\\hline
non paved-parking-place & impervious-surface\\\hline
bikeways & impervious-surface\\\hline
sidewalks & impervious-surface\\\hline
entrance-exit & impervious-surface\\\hline
danger-area & impervious-surface\\\hline
lane-markings & impervious-surface\\\hline
danger-area & impervious-surface\\\hline
car & vehicle\\\hline
trailer & clutter\\\hline
van & vehicle\\\hline
truck & vehicle\\\hline
large-truck & vehicle\\\hline
bus & vehicle\\\hline
clutter & clutter\\\hline
impervious-surface & impervious-surface\\\hline
tree & tree\\\hline
	\end{tabular}
	}\label{tab:mergeforpostdam}
\end{table}

\section{Further quantitative results}
\label{sec:supplementaryQuantitativeResults}
In~\cref{tab:bchmkSkyscapesAllcls}, we present an extensive 
benchmark on SkyScapes-Dense using several different methods ranging from the initial FCN8, as the first semantic segmentation method that uses fully convolutional neural networks, to the very recent DenseASPP, BiSeNet, and DeepLabv3+ algorithms. ~\Cref{tab:bchmkSkyscapesEachcls}  shows the $IoU_{class}$, \ie the IoU for each of the 20 classes separately. 
Similarly, ~\cref{tab:bchmkSkyscapesMultiLaneAllclsGeneral} and
~\cref{tab:bchmkSkyscapesMultiLaneEachcls} show the benchmark results on SkyScapes-Lane (overall and for each class separately). 
Finally, results for the merged dense classes (the SkyScapes-Dense-Category task) are given in ~\cref{tab:bchmkSkyscapesAllclsNew} and
~\cref{tab:bchmkSkyscapesEachclsNew}. 
\experimentalresultsSkyScapesNew
\experimentalresultsSkyScapesIoUsNew
\experimentalresultsSkyScapesMultiClassLaneNew
\experimentalresultsSkyScapesMultiClassLaneIoUsNew

\experimentalresultsSkyScapesCategoryNew
\experimentalresultsSkyScapesCategoryIoUsNew

\section{Further qualitative results}
We also provide more qualitative results to demonstrate the generalization capability of our method. \Cref{fig:wv4} shows the satellite image of the whole area of Munich, Germany. This image was taken by the WorldView4 satellite with a \gls{gls:GSD} of \SI{30}{cm}.

The patches in~\cref{fig:wv4patches} highlight binary lane-marking segmentation results on the satellite image, the feasibility of which is, to our knowledge, demonstrated here for the first time. In this work, we expanded the work of Azimi et al.~\cite{azimi2018aerial} on binary lane-marking extraction.
It is thus feasible to extract whole-city lane-marking maps from a single satellite image.

~\Cref{fig:3k1},~\cref{fig:3k2}, and ~\cref{fig:3k3} show further qualitative results on three aerial images with different scales, \gls{gls:GSD}, illumination conditions, and from different geographical areas. These figures show the whole-image dense prediction and zoomed-in sample areas with dense, multi-class lane-marking, and multi-class edge segmentations.


  
\clearpage
    \afterpage{%
\begin{longtable}[h]
{ m{3cm} m{3cm} m{5cm} c  }
\caption{List of categories including their definition and a typical example.}
    \hline
   Category & Class & Definition & Examples \\\hline 
  \endfirsthead
  \endhead
  \endfoot
  \endlastfoot
   nature & low vegetation & Includes all natural areas without large plants, \eg lawns.  & 
   \begin{minipage}{.3\textwidth}
   \vspace{0.1cm}
        \includegraphics[width=\linewidth]{figures/examples/new/lv.jpg}
    \end{minipage}
     \\ 
    & tree & Areas covered by large plants, such as trees or large bushes. & 
    \begin{minipage}{.3\textwidth}
    \vspace{0.1cm}
      \includegraphics[width=\linewidth]{figures/examples/new/tree.jpg}
      \vspace{-0.3cm}
    \end{minipage}
   \\ \hline
   residential & building  &  Structures with walls and a roof, such as houses, factories, and garages.  & 
    \begin{minipage}{.3\textwidth}
    \vspace{0.1cm}
        \includegraphics[width=\linewidth]{figures/examples/new/b.jpg}
      \vspace{-0.3cm}
    \end{minipage}
    \\ \hline
    vehicle area & paved-road & Includes all roads that are asphalted. &
    \begin{minipage}{.3\textwidth}
      \vspace{0.1cm}
      \includegraphics[width=\linewidth]{figures/examples/new/pr.jpg}
    \end{minipage}
    \\
    & non-paved-road & All roads that are not paved, \eg forest roads, dirt roads, and unsurfaced roads. &
    \begin{minipage}{.3\textwidth}
      \vspace{0.1cm}   
      \includegraphics[width=\linewidth]{figures/examples/new/npr.jpg}
    \end{minipage}
    \\
    & paved-parking-place & includes all asphalted areas for parking vehicles, such as car parks. The parking area include the vehicle as well which has not been shown in the figure &
    \begin{minipage}{.3\textwidth}
    \vspace{0.1cm}
      \includegraphics[width=\linewidth]{figures/examples/new/pp.jpg}
    \end{minipage}
    \\
    & non-paved-parking-place & Unsurfaced areas used for parking. The parking area include the vehicle as well which has not been shown in the figure. &
    \begin{minipage}{.3\textwidth}
    \vspace{0.1cm}
      \includegraphics[width=\linewidth]{figures/examples/new/npp.jpg}
      \vspace{-0.3cm}
    \end{minipage}
    \\ \hline
    
    lane-markings & long line & Thin solid lines, such as no passing lines or roadside markings. &
    \begin{minipage}{.3\textwidth}
    \vspace{0.1cm}
      \includegraphics[width=\linewidth]{figures/examples/new/ll.jpg}
    \end{minipage}
    \\
    & dash line & Any broken line with long line segments, \eg lane separators. &
    \begin{minipage}{.3\textwidth}
    \vspace{0.1cm}
      \includegraphics[width=\linewidth]{figures/examples/new/dl.jpg}
    \end{minipage}
    \\
    & tiny dash line & Any broken line with tiny line segments, \eg lines enclosing pedestrian crossings. &
        \begin{minipage}{.3\textwidth}
        \vspace{0.1cm}
      \includegraphics[width=\linewidth]{figures/examples/new/tdl.jpg}
    \end{minipage}
    \\
    & zebra zone & 
Areas with diagonal lines, \eg restricted zones. &
    \begin{minipage}{.3\textwidth}
    \vspace{0.1cm}
      \includegraphics[width=\linewidth]{figures/examples/new/zz.jpg}
    \end{minipage}
    \\
    & turn sign & Arrows on the road, such as intersection arrows or merge arrows. &
    \begin{minipage}{.3\textwidth}
    \vspace{0.1cm}
      \includegraphics[width=\linewidth]{figures/examples/new/ts.jpg}
    \end{minipage}        
    \\
    & stop line & Thick solid line across lanes that signal to stop behind the line. &
    \begin{minipage}{.3\textwidth}
    \vspace{0.1cm}
      \includegraphics[width=\linewidth]{figures/examples/new/sl.jpg}
    \end{minipage}        
        \\
    & parking zone & Includes any lines that mark parking spots. &
    \begin{minipage}{.3\textwidth}
    \vspace{0.1cm}
      \includegraphics[width=\linewidth]{figures/examples/new/pz.jpg}
    \end{minipage}        
    \\
    & no parking zone & Zig-zag lines next to the curb mark that indicate that stopping or parking is forbidden. &
    \begin{minipage}{.3\textwidth}
    \vspace{0.1cm}
      \includegraphics[width=\linewidth]{figures/examples/new/npz.jpg}
    \end{minipage}    
    \\
    & crosswalk & Zebra-striped markings across the roadway mark a pedestrian crosswalk. &
    \begin{minipage}{.3\textwidth}
    \vspace{0.1cm}
      \includegraphics[width=\linewidth]{figures/examples/new/cw.jpg}
    \end{minipage}        
    \\
    & plus sign & All crossing tiny lines. &
    \begin{minipage}{.3\textwidth}
    \vspace{0.1cm}
      \includegraphics[width=\linewidth]{figures/examples/new/ps.jpg}
    \end{minipage}    
    \\
    & other signs & Includes all other signs, \eg numbers that indicate the speed limit. &
    \begin{minipage}{.3\textwidth}
    \vspace{0.1cm}
      \includegraphics[width=\linewidth]{figures/examples/new/os.jpg}
    \end{minipage}        
    \\
    & rest of lane-markings & Any other lane-marking. &
    \begin{minipage}{.3\textwidth}
    \vspace{0.1cm}
      \includegraphics[width=\linewidth]{figures/examples/new/r.jpg}
      \vspace{-0.3cm}
    \end{minipage}    
    \\ \hline
    human area & sidewalk &  Path with a hard surface on one or both sides of a road for pedestrians. &
    \begin{minipage}{.3\textwidth}
    \vspace{0.1cm}
      \includegraphics[width=\linewidth]{figures/examples/new/sw.jpg}
    \end{minipage}     
    \\
     & bikeway & Includes all lanes or roads for bikes. &
    \begin{minipage}{.3\textwidth}
    \vspace{0.1cm}
      \includegraphics[width=\linewidth]{figures/examples/new/bw.jpg}
    \end{minipage}      
    \\
    & danger-area & The intersection of bikeways with road marked with red, blue or green in Germany and some other countries &
    \begin{minipage}{.3\textwidth}
    \vspace{0.1cm}
      \includegraphics[width=\linewidth]{figures/examples/new/da.jpg}
      \vspace{-0.3cm}
    \end{minipage}      
    \\ \hline
    shared area & entrance-exit & All entrance and exit areas that are shared with pedestrians.&
    \begin{minipage}{.3\textwidth}
        \vspace{0.1cm}
      \includegraphics[width=\linewidth]{figures/examples/new/ee.jpg}
      \vspace{-0.3cm}
    \end{minipage}      
    \\ \hline
    vehicle & car & Includes all cars except vans. &
    \begin{minipage}{.3\textwidth}
        \vspace{0.1cm}
      \includegraphics[width=\linewidth]{figures/examples/new/car.jpg}
    \end{minipage}      
    \\ 
    & van & Any vehicles with box-like shapes. &
     \begin{minipage}{.3\textwidth}
         \vspace{0.1cm}
      \includegraphics[width=\linewidth]{figures/examples/new/van.jpg}
    \end{minipage}     
    \\ 
    & truck& Includes all small trucks such as delivery trucks.  &
    \begin{minipage}{.3\textwidth}
        \vspace{0.1cm}
      \includegraphics[width=\linewidth]{figures/examples/new/tk.jpg}
    \end{minipage}      
    \\ 
    & long-truck & All long trucks such as heavy goods vehicles.   &
    \begin{minipage}{.3\textwidth}
        \vspace{0.1cm}
      \includegraphics[width=\linewidth]{figures/examples/new/lt.jpg}
    \end{minipage}      
    \\ 
    & trailer & Includes all trailers that can be attached to any vehicle, \eg trucks or cars. &
    \begin{minipage}{.3\textwidth}
        \vspace{0.1cm}
      \includegraphics[width=\linewidth]{figures/examples/new/tr.jpg}
    \end{minipage}      
    \\ 
    & bus & Any buses including tourist coaches, school buses, and public buses.  &
    \begin{minipage}{.3\textwidth}
        \vspace{0.1cm}
      \includegraphics[width=\linewidth]{figures/examples/new/bus.jpg}
      \vspace{-0.3cm}
    \end{minipage}      
    \\ \hline
    other & impervious surface & Includes all other surfaces, such as construction sites, and non-temporary obstacles road users cannot go through (\eg low wall, rocky terrain, river).&
    \begin{minipage}{.3\textwidth}
        \vspace{0.1cm}
      \includegraphics[width=\linewidth]{figures/examples/new/is.jpg}
    \end{minipage}      
    \\ 
    & clutter & Includes all other human made structures, such as garbage bins, fences, or outdoor furniture. &
    \begin{minipage}{.3\textwidth}
        \vspace{0.1cm}
      \includegraphics[width=\linewidth]{figures/examples/new/cl.jpg}
      \vspace{-0.3cm}
    \end{minipage}      
    \\ \hline

  \label{tab:examples}
  
\end{longtable}
    }
\clearpage

\wv
\wvpatches
\threeKone
\threeKtwo
\threeKthree